\documentclass[pdflatex,sn-mathphys-num]{sn-jnl}% Math and Physical Sciences Numbered Reference Style
\usepackage{enumitem}
\usepackage{graphicx}%
\usepackage{multirow}%
\usepackage{amsmath,amssymb,amsfonts}%
\usepackage{amsthm}%
\usepackage{mathrsfs}%
\usepackage{mathtools}
\usepackage[title]{appendix}%
\usepackage{xcolor}%
\usepackage{textcomp}%
\usepackage{manyfoot}%
\usepackage{booktabs}%
\usepackage{algorithm}%
\usepackage{algorithmicx}%
\usepackage{algpseudocode}%
\usepackage{listings}%
\usepackage{makecell}
\usepackage{float} 
\theoremstyle{thmstyleone}%
\theoremstyle{thmstyletwo}%

\theoremstyle{thmstylethree}%

\begin{document}

\title[Article Title]{A Survey on Progress in LLM Alignment from the Perspective of Reward Design}


\author[1,2]{\fnm{} \sur{Miaomiao Ji}}\email{jimiaomiao@stu.scu.edu.cn}

\author[1]{\fnm{} \sur{Yanqiu Wu}}\email{yanqiu.wu@mq.edu.au}

\author[2]{\fnm{} \sur{Zhibin Wu}}\email{zhibinwu@scu.edu.cn}

\author[3]{\fnm{} \sur{Shoujin  Wang}}\email{shoujin.wang@uts.edu.au}

\author[1]{\fnm{} \sur{Jian  Yang}}\email{jian.yang@mq.edu.au}

\author[1]{\fnm{} \sur{Mark  Dras}}\email{mark.dras@mq.edu.au}

\author*[1]{\fnm{} \sur{Usman Naseem}}\email{usman.naseem@mq.edu.au}

\affil[1]{\orgdiv{School of Computing}, \orgname{Macquarie University}, \orgaddress{\street{4 Research Park Drive}, \city{Sydney}, \postcode{2109}, \state{NSW}, \country{Australia}}}

\affil[2]{\orgdiv{Business School}, \orgname{Sichuan University}, \orgaddress{\street{No. 29, Wangjiang Road}, \city{Chengdu}, \postcode{610065}, \state{Sichuan}, \country{China}}}

\affil[3]{\orgdiv{Data Science Institute}, \orgname{University of Technology Sydney}, \orgaddress{\street{15 Broadway}, \city{Sydney}, \postcode{2007}, \state{NSW}, \country{Australia}}}

\abstract{Reward design plays a pivotal role in aligning large language models (LLMs) with human values, serving as the bridge between feedback signals and model optimization. This survey provides a structured organization of reward modeling and addresses three key aspects: mathematical formulation, construction practices, and interaction with optimization paradigms. Building on this, it develops a macro-level taxonomy that characterizes reward mechanisms along complementary dimensions, thereby offering both conceptual clarity and practical guidance for alignment research. The progression of LLM alignment can be understood as a continuous refinement of reward design strategies, with recent developments highlighting paradigm shifts from reinforcement learning (RL)-based to RL-free optimization and from single-task to multi-objective and complex settings.

}




\maketitle

\section{Introduction}
\label{1}
\subsection{Challenges faced by LLMs and LLMs alignment}
In recent years, \textbf{large language models} (LLMs) such as GPT-4~\cite{openai2023gpt4}, Claude~\cite{anthropic2023claude}, and Gemini~\cite{google2023gemini} have demonstrated remarkable capabilities across a wide range of natural language understanding and generation tasks. These systems, built upon transformer architectures~\cite{vaswani2017attention} and trained on massive corpora, exhibit strong performance in zero-shot and few-shot learning, enabling a wide range of applications, from education and healthcare to programming and research assistance~\cite{bommasani2021opportunities, wang2024large}. 

However, as LLMs continue to gain influence, there is an urgent need to ensure that these models behave in ways that are beneficial, safe, and aligned with human intentions. LLMs are now expected to follow the principles of being \textbf{helpful}, \textbf{harmless}, and \textbf{honest} (HHH) \cite{askell2021general}. 
Despite their impressive capabilities, real-world deployments of LLMs frequently suffer from a number of critical challenges: factual inaccuracies~\cite{ji2023survey}, harmful or toxic outputs~\cite{gehman2020realtoxicityprompts}, persistent biases~\cite{bender2021dangers}, and unpredictable behavior in dynamic contexts~\cite{bommasani2021opportunities}. For instance, LLMs often generate hallucinated or incorrect information, which seriously limits their use in high-stakes domains such as healthcare or law \cite{ji2023survey}. They may also produce toxic, offensive, or harmful content due to biases present in the training data or insufficient filtering mechanisms \cite{gehman2020realtoxicityprompts}. In addition, persistent bias and fairness issues arise in the form of gender, racial, or cultural stereotyping, often reflecting societal inequalities embedded in the pretraining corpora \cite{bender2021dangers}. Ethical concerns also intensify in cross-cultural deployments, where differing moral standards complicate the alignment of model behavior with human values \cite{gabriel2020artificial}. Furthermore, controllability and interpretability remain limited, making it difficult for users to understand or guide the model’s decision-making process in a predictable manner \cite{bommasani2021opportunities}. Finally, privacy and security risks are increasingly pressing, as models may memorize and disclose sensitive user data and are vulnerable to prompt injection and adversarial attacks \cite{carlini2023extracting}.
These issues not only undermine trust but also highlight the broader difficulty of aligning such models with complex, diverse, and evolving human values.

To address these concerns, the alignment of LLMs has become a cornerstone of safe and responsible AI research. Alignment, in the context of artificial intelligence, refers to the extent to which a model's behavior reflects human values, goals, and ethical norms~\cite{gabriel2020artificial}. Yet achieving alignment in practice requires not only model-level interventions but also a robust, scalable methodology for systematically incorporating human preferences into the optimization process. 

Among various alignment techniques, \textbf{Reinforcement Learning from Human Feedback} (RLHF) has emerged as one of the most widely adopted paradigms~\cite{christiano2017deep, ouyang2022training}. RLHF enables the incorporation of human preferences into model training by using a \textbf{reward model} (RM) to guide \textbf{reinforcement learning} (RL) optimization. An RM is an optimization target or evaluation signal during training  that quantifies the alignment between model outputs and human preferences or alignment objectives. Rather than relying solely on \textbf{supervised fine-tuning} (SFT), RLHF leverages ranked or pairwise human feedback to train an RM, which in turn evaluates and ranks candidate outputs. These scores are then used to optimize the base LLM via algorithms such as \textbf{Proximal Policy Optimization} (PPO)~\cite{schulman2017proximal}. Although RLHF  has achieved notable progress, their effectiveness is still constrained by several fundamental limitations. One major issue is the subjectivity and inconsistency of human feedback. Annotators may provide conflicting judgments due to personal beliefs or differing interpretations of context, which introduces noise into the reward modeling process and destabilizes optimization \cite{ouyang2022training}. Another challenge is the scarcity and high cost of high-quality feedback, particularly in expert-driven domains such as medicine and law \cite{ziegler2019fine}. In addition, current alignment methods are prone to mode collapse, where models tend to produce repetitive and homogeneous outputs as a result of over-optimization on narrow reward signals \cite{jaques2019way}. Related to this is the issue of reward hacking, where models learn to exploit flaws in the reward function to artificially maximize scores without actually aligning with the intended human goals \cite{amodei2016concrete}. Final instability remains a common problem in RL settings, which often struggle with convergence due to high-dimensional action spaces and non-stationary reward landscapes \cite{christiano2017deep}. Together, these challenges reveal the limitations of current alignment strategies and point to the need for more resilient, adaptable, and principled approaches. Within this context, reward design has become an essential component of LLM alignment. Although LLMs exhibit powerful generative capabilities, they fundamentally depend on external signals to guide their behavior. At the same time, RMs are themselves imperfect, and their limitations can unintentionally amplify undesirable behaviors in LLMs. As a result, effective and well-structured reward design is critical for ensuring that model outputs remain aligned with human values, intentions, and expectations.

\subsection{Reward modeling: A central solution to alignment challenges}
Reward design plays a pivotal role in bridging the gap between raw model capabilities and meaningful alignment with human values, acting as the central connective tissue in the LLM alignment pipeline. Situated between feedback collection and optimization, it serves not only as a transformation layer that distills raw or subjective human feedback into actionable signals, but also as a high-leverage point for encoding abstract human preferences, safety considerations, and societal norms into quantifiable objectives~\cite{bai2022training}. In doing so, it links human intentions with the optimization objectives of LLMs, effectively shaping model behavior in complex, high-dimensional environments where direct supervision is often infeasible~\cite{zheng2023secrets, ouyang2022training}.

At the heart of this process lies the design and training of the RM, which operationalizes alignment through the reward function. This function determines which behaviors are encouraged or penalized, thus playing a foundational role in ensuring that LLM outputs reflect desirable and aligned behavior. A well-crafted reward function can mitigate the ambiguity and inconsistency inherent in human feedback by synthesizing diverse data sources and aligning them with high-level goals in a form amenable to optimization.

Despite its central role in the alignment pipeline, reward design remains a fundamentally open and challenging problem. RMs, as surrogate objectives, are designed to approximate the true preferences and expectations of human users. However, the inherent complexity and subjectivity of human values, coupled with the diversity of tasks and application scenarios, make it nearly impossible to construct a reward function that fully captures these nuanced preferences. Poorly constructed RMs often lead to unintended and misaligned behaviors, a phenomenon commonly referred to as reward misspecification~\cite{skalse2022defining}. This misspecification arises from the discrepancy between the designed reward signal and the users’ latent preferences, introducing significant risks during model training. When optimization is guided by imperfect or distorted RMs, the model may inadvertently exploit spurious correlations or overfit to proxy metrics, thereby deviating from genuine human intent~\cite{amodei2016concrete}. Such misalignments can result in unforeseen consequences, where models achieve high scores on evaluation metrics yet fail to deliver outputs that truly satisfy user needs. The situation becomes even more critical when these deviations are obscured by existing benchmarks, creating a misleading illusion of optimal performance while masking substantial alignment failures. Moreover, RMs frequently struggle with generalization across domains, leading to distributional shifts that undermine alignment robustness in real-world deployment scenarios~\cite{gao2023scaling}. Learned RMs are also susceptible to inaccuracies in preference data and the introduction of annotator biases, which complicate efforts to ensure fairness, robustness, and interpretability. These challenges are further exacerbated in emerging contexts involving multi-modal inputs~\cite{tsimpoukelli2021multimodal}, multi-turn interactions~\cite{yao2023react}, and multi-task generalization~\cite{xu2023multiinstruct}, where human values are heterogeneous, dynamic, and often underspecified.

The complexity of reward design is further magnified in emerging contexts involving multi-modal inputs~\cite{tsimpoukelli2021multimodal}, multi-turn interactions~\cite{yao2023react}, and multi-task generalization~\cite{xu2023multiinstruct}, where human values may be heterogeneous, evolving, or underspecified. As such, reward design must move beyond static, hard-coded rules toward data-driven models that dynamically adapt to evolving user needs and social contexts.

Ultimately, reward design is not just a technical component of RLHF, but the linchpin. It directly influences the behavior of downstream optimization algorithms, whether based on RL, supervised preference modeling, or in-context adaptation, by defining what constitutes a better output in any given context. To fulfill this role effectively, reward design must strike a careful balance among expressiveness, interpretability, robustness, and generalizability. More than just ensuring behavioral mimicry, it enables principled generalization from finite human data to diverse, open-ended real-world scenarios. In anchoring alignment research in a direction that is both scientifically rigorous and socially responsive, reward design embodies the theoretical foundations and practical mechanisms essential for aligning LLMs with human-centered goals.

\textbf{Motivation}
Although the alignment of LLMs has seen remarkable progress, there remains a critical gap in the literature: a comprehensive and systematic investigation into the central role of reward design within alignment paradigms is still missing. Most existing studies approach alignment from a general perspective~\cite{shen2023large, wang2023aligning, wang2024comprehensive, jiang2024survey}, often treating RMs as supplementary components rather than foundational mechanisms. While some recent efforts~\cite{zhong2025comprehensive} have started to examine the taxonomy and challenges of reward modeling, these works remain largely descriptive, focusing on existing system architectures or application scenarios without capturing the methodological evolution of reward design itself.

In contrast, this review takes a fundamentally different approach: we position reward design as a methodological paradigm that both reflects and drives the evolution of LLM alignment techniques. Unlike conventional reward specification in traditional machine learning, reward design in the context of LLM alignment involves fundamentally different considerations, including how alignment objectives are defined, how feedback signals are represented, and how reward signals are incorporated into optimization procedures. As alignment techniques continue to evolve, these considerations have become critical to ensuring both performance and robustness.

Moreover, many of the core challenges in aligning LLMs can often be traced back to the strategies and mechanisms used in reward modeling. In this context, reward design not only serves as a guiding mechanism for model behavior, but also provides a valuable analytical lens for understanding the trajectory of alignment research. Analyzing the evolution of reward design allows for a more systematic interpretation of paradigm shifts in alignment strategies and facilitates the identification of emerging methodological trends.

\textbf{Contributions} Our contributions are summarized as follows.
\begin{itemize}
	
	\item \textbf{Structured organization of reward modeling:}  
	Based on existing research, this survey organizes the space of reward modeling for LLM alignment along multiple dimensions. It highlights three foundational aspects: (i) mathematical formulation of reward models, (ii) construction practices, and (iii) interactions with optimization paradigms. Building on this, a macro-level taxonomy is presented that categorizes reward mechanisms into rule-based, data-driven, and hybrid approaches, as well as numerical vs. non-numerical and explicit vs. implicit types (see Figure~\ref{fig1}). This organization provides conceptual clarity and practical guidance for analyzing, comparing, and applying reward modeling methods.
	
	\item \textbf{Comprehensive analysis of hybrid reward design:}  
	Existing studies on hybrid reward mechanisms are systematically reviewed, with emphasis on the diverse sources of reward signals and the strategies employed for their integration. This analysis consolidates fragmented approaches into a coherent design space, offering methodological foundations for managing preference diversity, resolving conflicting objectives, and enabling alignment in complex real-world scenarios.
	
	\item \textbf{Synthesis of paradigm shifts in reward modeling for LLM alignment:}  
	The evolution of reward modeling is examined, covering the shift from RL-based to RL-free methods and from single-task to multi-task and multi-modal contexts. Driven by practical needs such as concurrent objective optimization, cross-modal consistency, and heterogeneous preferences, these shifts have stimulated continuous innovation in reward mechanism design. Treating the reward function as a lever for improving alignment efficiency highlights a promising direction for advancing LLM performance.
	
\end{itemize}
\begin{figure*}[htbp]
	
	\centering
	\includegraphics[width=\textwidth]{F1.pdf} 
	\caption{\label{fig1} 
		A hierarchical taxonomy of reward modeling in LLM alignment. 
	}
\end{figure*}

The remainder of this paper is organized as follows. Section~\ref{2} presents a diagnosis–prescription–treatment–inspired conceptual framework for LLM alignment, visually capturing the key processes and essential components involved, with a particular focus on reward modeling as a central solution to alignment challenges. Sections~\ref{3}–\ref{5} examine RM design from three complementary perspectives: mathematical formulation, methodological modeling and construction, and functional roles under different optimization paradigms. Building on these perspectives, a high-level categorization framework is introduced, covering: (i) Numerical vs.\ Non-numerical RMs; (ii) Rule-based, Data-driven, and Hybrid RMs; and (iii) Explicit vs.\ Implicit RMs. Section~\ref{6} analyzes recent methodological trends in reward design and their impact on the evolving landscape of LLM alignment, highlighting the shift from RL-based to RL-free approaches and the increasing demand for methods that address multi-objective, multi-task, and multi-modal scenarios. Promising future directions for advancing reward design are also outlined. Finally, Section~\ref{7} concludes the paper.

\section{Conceptual framework}
\label{2}
 LLM alignment can guide failure modes toward generating outputs that are consistently HHH~\cite{askell2021general}. To elucidate the critical role of reward design in achieving effective LLM alignment, it is helpful to begin with a structured conceptual framework that captures the core components and their interactions. Drawing inspiration from~\cite{pan2024automatically},  a medical metaphor is employed to illustrate the full pipeline from input to output, comprising the stages of feedback (diagnosis), reward design  (prescription), and optimization (treatment).

\begin{figure*}[htbp]
	\centering
	\includegraphics[width=0.7\textwidth]{F2.pdf} 
	\caption{\label{fig2} 
		A novel conceptual framework views LLM alignment as a medical treatment process. In this analogy, the model (the patient) produces raw outputs, which are then evaluated (diagnosed) via human or automated feedback. Based on this diagnosis, a tailored reward function (prescription) is crafted to guide the model’s optimization (treatment), ensuring that its behavior improves toward desired alignment goals (clinical targets). This analogy emphasizes the central role of reward design as the link between observation (feedback) and intervention (optimization), situating it at the core of the alignment pipeline. 
	}
\end{figure*}

\noindent\textbf{Feedback (diagnosis):} An LLM \( M : \mathcal{X} \to \mathcal{Y} \) performs a specific task by mapping an input \( x \in \mathcal{X} \) to an output text \( \hat{y} \in \mathcal{Y} \). A feedback model \( F : \mathcal{X} \times \mathcal{Y} \to \mathcal{Z} \) provides structured information about the quality or properties of a response. Given an input \( x \in \mathcal{X} \) and a response \( \hat{y} \in \mathcal{Y} \), which may come from a model, a human, or another agent, it computes \( z = F(x, \hat{y}) \in \mathcal{Z} \), where \( \mathcal{Z} \) captures diagnostic signals such as error types, critiques, explanations, or evaluations that help interpret or improve the response. During this stage, the model’s outputs are evaluated for any misalignments or undesired behaviors. Feedback plays a crucial role in identifying areas where the model deviates from its intended goals or ethical considerations. Feedback can take various forms, such as human-generated feedback, AI-generated feedback, binary feedback, preference feedback, pairwise feedback, and listwise feedback, which collectively inform the reward design process.

\noindent\textbf{Reward design (prescription):} An RM \( R : \mathcal{X} \times \mathcal{Y} \to \mathbb{R} \) assigns a scalar score to a model output. Given an input \( x \in \mathcal{X} \) and a response \( \hat{y} \sim M(x) \), it computes \( r = R(x, \hat{y}) \in \mathbb{R} \), where higher \( r \) indicates better alignment with human preferences. This is a pivotal stage where a reward function is crafted to guide the model’s behavior toward desired directions. Reward design provides structured incentives that steer the model toward generating outputs aligned with human values, ensuring that the content is safer and more ethical.

\noindent\textbf{Optimization (treatment):} Given a distribution over inputs \( x \in \mathcal{X} \) and an RM \( R : \mathcal{X} \times \mathcal{Y} \to \mathbb{R} \), the optimization goal is to adjust the pretrained model \( M \) that generates responses \( \hat{y} \sim M(x) \) which maximize expected reward,
$
\mathbb{E}_{x, \hat{y} \sim M(x)}[R(x, \hat{y})]$.
In this stage, the model undergoes iterative refinement using optimization techniques, such as supervised learning (SL), RL, or ICL. The goal is to enhance the model’s performance and alignment, ensuring it generates high-quality outputs that adhere to ethical and societal standards.

This framework also provides important insights for reward design: reward modeling in LLM alignment involves balancing multi-dimensional feedback and managing trade-offs among multiple objectives, while maintaining a task-oriented focus. Although multi-dimensional reward design is desirable, effective strategies should not aim for superficial comprehensiveness. Instead, they should adopt a task-driven, symptom-targeted approach, formulating precise reward strategies that address specific alignment deficiencies. Furthermore, while achieving high-quality outputs remains the ultimate goal, the stability and robustness of the optimization process are equally critical considerations. By incorporating intermediate optimization indicators and process-level feedback into reward design, the training process can achieve greater controllability and reliability, thereby reducing the risks of divergence, overfitting, and pathological behaviors, and ultimately enabling more robust and efficient model alignment.

\section{Reward model: mathematical formulation}
\label{3}
\subsection{Numerical reward modeling}
This section explores how variations in learning paradigms, input data formats, information granularity, and output forms drive the evolution and adaptation of the mathematical formulation of	RMs.

\subsubsection{Pointwise reward modeling and preferencewise modeling}

A standard way to construct an RM is to regress a scalar function
\(r_{\phi}(x,y)\) onto absolute human scores.
Given a labeled corpus
\(\mathcal{D}=\{(x^{(i)},y^{(i)},s^{(i)})\}_{i=1}^{N}\),
where \(x^{(i)}\) is the prompt, \(y^{(i)}\) the system response and
\(s^{(i)}\in\mathbb{R}\) a human quality score
(e.g.\ a Likert value in \([0,1]\) or on a 1–5 scale),
the model is trained by minimizing mean-squared error
\begin{equation}
\label{EQ1}
\mathcal{L}_{\text{MSE}}(\phi)=
\frac1N\sum_{i=1}^{N}\bigl(r_{\phi}(x^{(i)},y^{(i)})-s^{(i)}\bigr)^{2}.
\end{equation}
This formulation treats the RM as a continuous regressor rather than a relative ranker. Despite providing explicit supervision, supervised scoring is susceptible to subjectivity and inconsistency in absolute ratings, and may fail to capture the full diversity of possible responses.

In the context of LLM alignment, reward modeling typically involves transforming human preference data into explicit scoring functions for optimization, as first introduced by \cite{christiano2017deep}. In most implementations,  RMs share the same Transformer backbone as the underlying language model, with the hidden representation of the final token projected through a linear layer to produce a scalar reward score \cite{jiang2024survey}. To enable a systematic analysis of reward modeling approaches, two key dimensions are commonly considered: the structure of preference feedback (e.g., pairwise vs. listwise), and the output format of the RM (e.g., single-value vs. multi-value). Each dimension is described in detail below.

Given a dataset of human preference annotations $\mathcal{D} = \{(x^{(i)},y_w^{(i)},y_\ell^{(i)})\}_{i=1}^N$, where for each prompt $x$ the response $y_w$ is preferred to $y_\ell$, a scalar RM $r_\phi(x,y)$ is trained by modeling the probability of the preferred response under the Bradley–Terry formulation \cite{bradley1952rank}:

\begin{equation}
\label{EQ2}
p^*\bigl(y_w \succ y_\ell \mid x\bigr)
= \frac{\exp\bigl(r_\phi(x,y_w)\bigr)}
{\exp\bigl(r_\phi(x,y_w)\bigr) + \exp\bigl(r_\phi(x,y_\ell)\bigr)}
= \sigma\!\bigl(r_\phi(x,y_w) - r_\phi(x,y_\ell)\bigr)
\end{equation}
Taking the logarithm gives the log-probability $\log p^*(y_w \succ y_\ell \mid x) = \log \sigma\!\bigl(r_\phi(x,y_w) - r_\phi(x,y_\ell)\bigr)$. The parameters $\phi$ are optimized by minimizing the negative log-likelihood over $\mathcal{D}$, yielding the loss:

\begin{equation}
\label{EQ3}
\mathcal{L}_{pairwise}\bigl(r_\phi,\mathcal{D}\bigr)
= -\mathbb{E}_{(x,y_w,y_\ell)\sim\mathcal{D}}
\!\Bigl[\log \sigma\bigl(r_\phi(x,y_w)-r_\phi(x,y_\ell)\bigr)\Bigr].
\end{equation}
Minimizing $\mathcal{L}_{pairwise}$ encourages the reward gap $r_\phi(x,y_w)-r_\phi(x,y_\ell)$ to be large whenever humans prefer $y_w$, thereby aligning the model’s scores with human judgments. 

\subsubsection{Pairwise preference modeling and listwise preference modeling}
Given a dataset of human ranking annotations $\mathcal{D} = \{(x^{(i)},\langle y_1^{(i)},\dots,y_K^{(i)}\rangle)\}_{i=1}^N$, where for each prompt $x$ the responses $\{y_j\}_{j=1}^K$ are ordered from most to least preferred, a scalar RM $r_\phi(x,y)$ is trained by modeling the probability of the observed ranking under a Plackett–Luce model \cite{plackett1975analysis}:
\begin{equation}
\label{EQ4}
p^*(y_1\succ \cdots \succ y_K \mid x)
= \prod_{j=1}^K 
\frac{\exp\bigl(r_\phi(x,y_j)\bigr)}
{\sum_{t=j}^K \exp\bigl(r_\phi(x,y_t)\bigr)}.
\end{equation}
Taking the logarithm yields the log-probability $\log p^*(y_1\succ \cdots \succ y_K \mid x) = \sum_{j=1}^K \bigl(r_\phi(x,y_j) - \log \sum_{t=j}^K \exp(r_\phi(x,y_t))\bigr)$. The parameters $\phi$ are optimized by minimizing the negative log-likelihood over $\mathcal{D}$, giving the loss
\begin{equation}
\label{EQ5}
\mathcal{L}_{listwise}(r_\phi,\mathcal{D})
= -\mathbb{E}_{(x,\mathbf{y})\sim\mathcal{D}}
\Bigl[\sum_{j=1}^K\bigl(r_\phi(x,y_j)
- \log \sum_{t=j}^K \exp\bigl(r_\phi(x,y_t)\bigr)\bigr)\Bigr].
\end{equation}
Minimization of $\mathcal{L}_{listwise}$ drives the reward differences between higher-ranked and lower-ranked responses to be large, thus aligning the model’s scores with full ranking judgments.

\subsubsection{Response-level reward modeling and token-level reward modeling}

Conventional RMs typically score entire responses, resulting in sparse feedback that may lead to unstable training and insufficient supervision of local quality. To address this, token-level RMs have been proposed to assign individual rewards to each token, offering a more detailed optimization signal. 
Given a dataset of human preference annotations
\(\mathcal{D} = \{(x^{(i)}, y_w^{(i)}, y_\ell^{(i)})\}_{i=1}^N\),
where each pair \((y_w, y_\ell)\) represents a preferred and less-preferred response to the same prompt \(x\), token-level reward modeling assigns a scalar reward to each token \(y_t\) within a response sequence \(y = (y_1, \dots, y_T)\), conditioned on its left context and the prompt. Formally, a token-level RM outputs a vector of token-wise scores:
$
\mathbf{q}_\phi(x, y) = \left(q_\phi(x, y_{\le1}),\, q_\phi(x, y_{\le2}),\, \dots,\, q_\phi(x, y_{\le T})\right),
$
where each \(q_\phi(x, y_{\le t}) \in \mathbb{R}\) denotes the predicted reward for the token \(y_t\), given the prefix \((y_1, \dots, y_{t-1})\) and the prompt \(x\).

To align token-level scores with human preferences, the aggregated sequence reward is defined by summing the token-wise scores:
$
R_\phi(x, y) = \sum_{t=1}^{T} q_\phi(x, y_t \mid y_{<t}),
$
and is used to model the pairwise preference probability under the Bradley–Terry formulation:
$
p^*\bigl(y_w \succ y_\ell \mid x\bigr)
= \sigma\bigl(R_\phi(x, y_w) - R_\phi(x, y_\ell)\bigr),
$
where \(\sigma(\cdot)\) denotes the sigmoid function.

Taking the logarithm yields the log-probability:
$
\log p^*\bigl(y_w \succ y_\ell \mid x\bigr) = \log \sigma\bigl(R_\phi(x, y_w) - R_\phi(x, y_\ell)\bigr),
$
and the corresponding training objective is defined as:
\begin{equation}
\label{EQ6}
\mathcal{L}_{\text{token-agg}}\bigl(q_\phi, \mathcal{D}\bigr)
= - \mathbb{E}_{(x, y_w, y_\ell) \sim \mathcal{D}} \left[
\log \sigma\bigl(R_\phi(x, y_w) - R_\phi(x, y_\ell)\bigr)
\right].
\end{equation}
Minimizing \(\mathcal{L}_{\text{token-agg}}\) encourages the cumulative reward of the preferred response \(y_w\) to exceed that of the less-preferred response \(y_\ell\), thus indirectly shaping token-level scores to reflect the quality contributions of individual tokens.

An alternative step-wise loss directly compares token-level scores:
\begin{equation}
\label{EQ7}
\mathcal{L}_{\text{token-step}}(q_\phi,\mathcal{D})=
-\mathbb{E}_{(x,y_w,y_\ell)\sim\mathcal{D}}
\left[\sum_{t=1}^{\max(T_w,T_\ell)}
\log\sigma\!\bigl(q_\phi(x,y_{w,t})-q_\phi(x,y_{\ell,t})\bigr)\right].
\end{equation}

Building on these baselines, token-level reward modeling has been advanced and broadened to accommodate diverse application needs. \citet{yoon2024tlcr}  introduced \textbf{Token-Level Continuous Reward} (TLCR) for RLHF, which used a discriminator to distinguish positive and negative tokens and assigned context-aware continuous rewards based on its confidence, achieving consistent improvements over sequence-level and token-level discrete rewards. \citet{xu2024finegrained} enhanced LLM alignment through fine-grained token-level supervision by asking annotators to minimally edit less preferred responses to create a refined dataset, which was used to train a token-level reward model and guide fine-grained PPO training.
\citet{zeng2024tdpo} introduced \textbf{Token-level Direct Preference Optimization} (TDPO), which utilized the Bradley–Terry model for a token-based reward system to enhance KL divergence regulation while preserving simplicity without explicit reward modeling. \citet{fu2025tldr} introduced a \textbf{Token-Level Detective Reward Model} (TLDR) to provide fine-grained annotations for large vision language models, using a perturbation-based method to generate synthetic hard negatives with token-level labels for training. \citet{chen2025qrm} proposed the \textbf{Q-function Reward Model} (Q-RM) by decoupling reward modeling from language generation and optimizing a discriminative policy for token-level rewards. Overall, token-level reward modeling provides finer credit assignment and more stable training, improving safety, controllability, and interpretability, and better supporting real-world deployment.

\subsubsection{Single-value reward modeling and multi-value reward modeling }

Depending on the output information density of the numerical preference models, RMs can be categorized as either single-value RMs or multi-value RMs. Single-value RMs assign a scalar score to each candidate output: $r_\phi(x, y) \in \mathbb{R}$, where \( x \) denotes the input (e.g., a prompt or instruction), \( y \) is the model-generated response, and \( r_\phi \) is a neural network parameterized by \( \phi \) that produces a scalar utility score. Multi-value RMs generate a vector of scores to capture multiple quality dimensions simultaneously:
$\mathbf{r}_\phi(x, y) = \left[ r_\phi^{(1)}(x, y),\, r_\phi^{(2)}(x, y),\, \dots,\, r_\phi^{(K)}(x, y) \right] \in \mathbb{R}^K$,
where each component \( r_\phi^{(k)}(x, y) \) corresponds to a specific evaluation objective, such as helpfulness, harmlessness, or honesty. This formulation supports multi-objective alignment and enables fine-grained supervision from human feedback.

Although single-value reward modeling already shows good performance in the aforementioned studies, there exist scenarios where its capability becomes limited. For instance, a single-value RM might overfit on the feedback data, have difficulty addressing multiple aspects of human preferences, or fall short in delivering appropriate preference signals for fine-grained portions of output sequences \cite{jiang2024survey}. To overcome these limitations, multi-value reward modeling approaches have been introduced. As an example, \citet{christiano2017deep} addressed overfitting by training multiple RMs on different random partitions of the preference dataset and averaging their individually normalized outputs to generate the final reward. Likewise, \citet{coste2023reward} reduced overfitting by combining models initialized with various random seeds, either by taking the minimum reward or applying a weighted penalty based on variance. In order to generate fine-grained reward signals, many works adopt multi-value reward modeling as a means to support alignment with multiple objectives in complex tasks. Rather than assigning individual scalar rewards to each objective, vectorized RMs embed the interdependencies among several quality metrics into a unified multidimensional format, which enables more coherent trade-offs and facilitates more efficient optimization across related goals. \citet{liu2023aligning} utilized activation vectors extracted from LLMs fine-tuned on preferred versus non-preferred outputs as predictors of rewards within the model's representational space. In addition, \citet{frans2024unsupervised} investigated how to encode latent reward functions into vectors derived from corresponding data samples, thereby enabling RL to operate across a variety of tasks using more flexible and generalizable reward signals.
\subsection{Non-numerical reward modeling }

Beyond numerical RMs, reward mechanisms that provide natural-language preference signals have also been explored. For example, \citet{wang2023shepherd} and \citet{cui2023ultrafeedback} introduced critic networks that learn from human critique text to evaluate model outputs and offer corrective suggestions, producing detailed assessments in prose form \cite{richardson2023syndicom}. The tool-augmented validation framework of \citet{li2023tool} likewise leveraged external utilities to generate natural‐language feedback as part of the reward signal. Moreover, \citet{akyurek2023rl4f} employed RL to train a critique‐generation module, using similarity metrics between human‐preferred outputs and those refined via generated critiques as the optimization signal. Some systems even interleave feedback within the generation process: for instance, Self-Refine \cite{madaan2023self} issued stepwise commentary during chain-of-thought reasoning to guide each subsequent generation rather than relying solely on a final aggregated score.   Non-numerical RMs convey discrete labels or natural-language feedback that must be interpreted or mapped before optimization, delivering richer contextual guidance, improved interpretability, and enhanced human-in-the-loop refinement.

\section{Reward model: methodological modeling and construction}
\label{4}
The construction paradigm of RMs constitutes a foundational aspect of LLM alignment. It critically shapes the model’s ability to capture and operationalize human preferences or normative objectives, while also influencing key properties such as generalization capacity, robustness, and training efficiency. Different paradigms, including rule-based, data-driven, and hybrid approaches, involve distinct trade-offs in terms of scalability, feedback dependency, interpretability, and adaptability to specific tasks. Therefore, the principled selection and design of construction methods are essential to ensure that RMs provide reliable and effective guidance for optimization toward aligned behavior. Table~\ref{table 2} summarizes the construction paradigms for RMs and their key characteristics.

\begin{table}[htbp]
	\caption{Contrasting functional roles of reward models under different alignment optimization paradigms}
	\label{table 2}
	\centering
	\footnotesize
	\begin{tabular}{
			>{\raggedright\arraybackslash}p{1.5cm}
			>{\raggedright\arraybackslash}p{3cm}
			>{\raggedright\arraybackslash}p{3cm}
			>{\raggedright\arraybackslash}p{4.2cm}
		}
		\toprule
		\textbf{Aspect} & \textbf{Rule-based RM} & \textbf{Data-driven RM} & \textbf{Hybrid RM} \\
		\midrule
		
		\textbf{Construction Approach} &
		Rewards are explicitly defined using handcrafted rules, external knowledge bases, or predefined heuristics. &
		Rewards are learned from labeled datasets through supervised learning, preference modeling, or inverse reinforcement learning. &
		Combines multiple reward signals (e.g., rule-based + data-driven), often through weighted fusion or multi-stage pipelines. \\
		\addlinespace
		
		\textbf{Strengths} &
		- High interpretability and transparency \newline
		- Strong controllability in safety-critical tasks &
		- Better scalability to complex tasks \newline
		- Can generalize from data beyond human-written rules &
		- Balances interpretability and expressiveness \newline
		- Increases robustness to reward misspecification \newline
		- Adapts to diverse alignment objectives \\
		\addlinespace
		
		\textbf{Limitations} &
		- Limited coverage and flexibility \newline
		- Difficult to scale in open-ended tasks \newline
		- Prone to rule misspecification &
		- Requires large, high-quality labeled data \newline
		- May overfit to annotation biases \newline
		- Harder to interpret learned rewards &
		- Increases system complexity \newline
		- Requires careful signal balancing and integration design \newline
		- Fusion strategies may introduce alignment trade-offs \\
		\addlinespace
		
		\textbf{Typical Examples} &
		- Toxicity filters \newline
		- Safety constraints via Wikidata queries \newline
		- Syntax/logic rule checkers &
		- Pairwise preference learning \newline
		- Reward modeling via classification/regression \newline
		- Inverse reinforcement learning &

		- Multi-source RM (e.g., rule-based + data-driven)
		
		- Multi-granularity RM (e.g., response-level + sentence-level + token-level)
		
		- Multi-aspect RM (e.g., fluency + factuality + safety) 
		
		- Multi-stage RM (e.g., training-time supervision + inference-time post-hoc scoring)
		
		- Multi-modal RM (e.g., text + images + audio) 
		
		 \\

		\addlinespace
		
		\textbf{Best Suited Scenarios} &
		- Safety-critical applications \newline
		- Tasks requiring strict compliance to explicit norms &
		- Tasks where human preference data is abundant \newline
		- Scenarios demanding nuanced semantic understanding &
		- Complex alignment tasks where single-source rewards are insufficient \newline
		- Contexts with competing alignment objectives (e.g., quality vs. safety) \\
		
		\bottomrule
	\end{tabular}
\end{table}

\subsection{Rule-based reward modeling and data-driven reward modeling}

In the development of LLM alignment, reward modeling has followed two primary technical trajectories: rule-based design schemes and data-driven learning paradigms.

Rule-based RMs were introduced in the early stages, focusing on the direct encoding of behavioral constraints into the system. These approaches relied on human experts to define explicit rule sets, offering high interpretability and precise control. Such characteristics made them particularly suitable for high-stakes scenarios where well-defined behavioral boundaries were essential. For example, \citet{mu2024rule} proposed the Rule-Based Rewards (RBR) framework, which combined manually specified behavioral rules with verification from a large language model. This framework enabled effective alignment without relying on large-scale human preference data. In addition to reducing dependence on manual annotation, it allowed for rapid updates in response to evolving safety requirements. Rule-based methods demonstrated particular robustness in environments where generalization was unreliable or preference data was limited. By explicitly constraining the model’s action space, these approaches also helped mitigate issues commonly observed in RLHF pipelines, such as reward hacking and overly conservative refusals.

In contrast, data-driven reward modeling did not initially focus on preference learning. Early methods primarily employed conventional classification or regression frameworks, where automated evaluator were used to assign quality scores to generated outputs. These scores served as training signals for learning reward functions. For instance, \citet{chang2025bleuberi} introduced BLEUBERI, a method that leveraged BLEU as a simple yet effective reward signal for aligning models to follow instructions. Similarly, \citet{xue2023improving} proposed RLFC, which computed reward values based on the factual consistency between generated responses and gold-standard knowledge, as assessed by \textbf{natural language inference} (NLI) models or QA-style matching systems. These signals guided models toward producing more factually accurate outputs through RL.

As model capabilities improved and large-scale human feedback became more readily available, the field gradually shifted toward preference learning, which is based on human-annotated rankings over multiple outputs, allowing models to learn implicit reward functions that better reflected nuanced user preferences. This approach became central to RLHF pipelines such as InstructGPT \cite{ouyang2022training}. Despite its expressive power, preference learning proved highly sensitive to annotation quality and was often plagued by inconsistencies and subjective disagreements in human judgments. Moreover, collecting high-quality, fine-grained preference data at scale posed significant cost and feasibility challenges.

To address these limitations, recent research has explored imitation-based alternatives, most notably \textbf{inverse reinforcement learning (IRL)}. IRL aims to infer latent reward functions from expert demonstrations, thereby avoiding the need for explicit scoring or preference annotation. This reoriented the objective of reward modeling from selecting the best output to understanding the underlying rationale behind expert behavior. For instance, \citet{sun2024inverse} introduced a framework based on trajectory distribution alignment to recover implicit reward structures. \citet{cai2024approximated} further advanced this line of work by proposing a variational Bayesian IRL objective to improve generalization across diverse task settings. Building upon these ideas, \citet{cheng2025dynamic} incorporated category-specific reward scaling into the IRL framework, enabling dynamic modulation of reward signals in response to contextual safety demands—an especially pertinent development for high-risk applications.

This progression reflects a fundamental shift in the objectives of reward modeling: from enforcing fixed behavioral rules to uncovering the latent structures of human values. The transition from rule specification to preference learning, and ultimately to behaviorally grounded reward inference, reveals a deepening understanding of alignment not merely as behavioral conformity, but as the construction of adaptable, value-sensitive feedback systems capable of responding to diverse and evolving alignment contexts.

\subsection{Hybrid reward modeling}

Human preferences are highly diverse and context-dependent, varying significantly across tasks, demographics, cultures, and individual users. At the same time, different tasks impose distinct criteria for what constitutes high-quality output, and real-world applications involve a wide range of objectives. The combination of preference diversity, task variability, and objective heterogeneity makes it extremely challenging to design reward functions that can fully capture such nuanced expectations. Consequently, static, universal RMs are often impractical for aligning model behavior with the complex and variable nature of human values. Furthermore, hand-engineered metrics can conflict with each other. For example, generating shorter, more precise descriptions may improve BLEU \cite{papineni2002bleu} scores while reducing ROUGE \cite{lin2003rouge} scores due to lower recall. These trade-offs are often task-specific and user-dependent, revealing the limitations and fragility of static, one-size-fits-all reward designs.

To address the limitations of RMs in terms of adaptability and expressiveness, researchers have proposed hybrid RMs. By devising targeted fusion strategies, hybrid RMs flexibly integrate complementary supervision signals across different granularities, sources, and modalities, enabling more efficient preference modeling and enhanced representational capacity. In complex tasks, multi-objective evaluation scenarios, or environments with noisy feedback, hybrid reward mechanisms have demonstrated superior robustness and generalization capabilities.

\subsubsection{Sources for hybrid reward modeling }

Human preferences are subjective, diverse, and context-dependent. Hybrid RMs allow for the integration of multiple supervision signals from different sources (e.g., human feedback, expert rules, user behavior, or model predictions), enabling better alignment with actual human preferences. \citet{peng2025agentic}  proposed agentic reward modeling, a reward system that combined RMs with verifiable correctness signals from multiple aspects to provide reliable supervision. They implemented a reward agent, RewardAgent, which combined human preference rewards with two verifiable signals: factuality and instruction following, to enhance the reliability of the reward signals.

In complex tasks or environments with noisy feedback, single reward signals may lead to overfitting to noise or biases. Multi-source rewards introduce redundancy by incorporating signals from diverse sources, which enhances the model’s robustness to anomalies and its generalization across varied inputs.
\citet{wang2025autorule} proposed AutoRule, a fully automated framework that extracted symbolic rules from preference feedback. These rules were used to compute an auxiliary reward signal that complemented the learned RM, resulting in improved RLHF performance and reduced reward hacking.

A single reward signal often covers only one aspect of a task, whereas hybrid rewards can combine signals across multiple objective dimensions to more comprehensively define what constitutes a “good” output. For example, in dialogue generation, the model needs to produce responses that are both fluent and factually accurate. In recommendation systems, it should match user interests while also maintaining diversity and novelty.
\citet{zhang2025dpa}  introduced the Directional Preference Alignment (DPA) framework to overcome the limitations of scalar-reward RLHF in representing diverse user preferences. DPA incorporated multi-objective reward modeling and modeled user preferences as unit vectors in reward space, enabling user-dependent control over generation behavior. They fine-tuned LLMs using a preference-conditioned variant of \textbf{Rejection Sampling Finetuning} (RSF), which achieved improved performance trade-offs across various reward objectives. \citet{zhang2025moslim} proposed MOSLIM, employed a multi-head RM to classify question-answer pairs and mapped these classifications into scalar rewards to to handle diverse objectives, enabling flexible control via prompting without requiring preference-specific training during the SFT phase. 

In real-world applications, evaluation metrics often exhibit trade-offs. For instance, increasing precision may reduce recall, generating concise outputs can compromise informational richness, and prioritizing safety may constrain creativity. Hybrid reward mechanisms offer a principled approach to balancing such competing objectives, enabling multi-objective optimization without sacrificing one goal for another.
Considering that RL-based fine-tuning was unstable and resource-heavy, especially under diverse and conflicting objectives, \citet{zhou2023modpo} proposed \textbf{Multi-Objective Direct Preference Optimization} (MODPO) as a RL–free alternative to DPO for multi-objective alignment. MODPO folded language modeling directly into reward modeling, enabling language models to act as implicit collective RMs.

Some tasks involve inputs from multiple modalities, such as text, images, audio, or video, or require optimization across both local and global levels. Hybrid reward mechanisms can combine signals across different modalities, granularities, and stages, making them suitable for more complex and realistic application scenarios.
\citet{sun2024aligning} proposed Factually Augmented RLHF, an approach that augmented the RM with additional factual information such as image captions and ground-truth multiple-choice options to mitigate the reward hacking problem.
Multi-granularity hybrid rewards focus on integrating alignment signals across varying levels of granularity, from fine-grained token-level correctness to high-level discourse structures. This paradigm is crucial for aligning LLMs on tasks where coherence, style, and factuality must be maintained across local and global scopes.  \citet{liu2025hafrm} proposed HAF-RM, a hybrid alignment framework for RM training that introduces an additional constraint on token-level policy probabilities alongside the conventional reward score. This framework enabled simultaneous supervision of the internal preference model at the token level while optimizing the reward mapping layer at the sequence level.  \citet{yu2024reasoning} argued that process supervision relied on learned reward models requiring costly data and suffered reward misalignment, while outcome supervision failed on complex multi-step tasks. They proposed Outcome Refining Process Supervision, which unified process and outcome supervision via executable verification and tree search to refine reward signals.

\subsubsection{Strategies for hybrid reward modeling}
Fusion mechanisms played a central role in hybrid reward modeling, determining how diverse reward signals were integrated into a unified optimization objective. These mechanisms were not merely technical components but strategic decisions that directly impacted a model’s ability to synthesize multidimensional feedback for achieving stable and trustworthy alignment.

One of the most commonly adopted approaches was weighted averaging, where reward signals were either combined using static weights or dynamically adjusted based on confidence, task context, or domain-specific attributes. While static weighting offered simplicity, it lacked the flexibility to handle conflicting or evolving alignment objectives.
To address this, \citet{liu2025amopo} proposed \textbf{Adaptive Multi-objective Preference Optimization} (AMoPO), which treated dimension-aware generation metrics as implicit rewards within a multi-objective optimization paradigm. They further introduced an adaptive weight assignment mechanism, where the generation space was modeled as a Gaussian distribution to enable dynamic prioritization across different preference dimensions.

In addition to averaging, researchers also explored the temporal structure of reward fusion. Fusion often occurred sequentially, with rule-based signals used for initial filtering or scaffolding, followed by refinement through learned or adaptive RMs. This reflected a coarse-to-fine alignment strategy that became increasingly prominent in LLM alignment research.
For example, \citet{bai2022constitutional} implemented a hybrid reward modeling framework by applying rule-based constitutional principles to remove unsafe responses before collecting AI feedback. The resulting RM, trained on revised and critiqued outputs, combined rule-based safety enforcement with preference alignment, improving harmlessness without relying solely on human labels.

Other studies introduced modular or hierarchical fusion strategies that enabled dynamic reward selection. These approaches employed separate discriminator or controller modules to choose appropriate reward signals based on task requirements or contextual information.
\citet{lai2025alarm} proposed ALaRM, a hierarchical reward modeling framework for RLHF. It decomposed alignment into holistic and aspect-specific sub-objectives. They introduced a consistency-based aggregation strategy to filter and combine multiple feedback signals. Their two-stage training process prioritized holistic rewards and incorporated aspect-specific feedback only when necessary to guide models toward better alignment.

Later work explored neural fusion mechanisms such as attention-based and gating-based approaches. These methods adaptively routed reward signals according to the input context and supported more fine-grained fusion across tasks or reward types.
\citet{qiu2024sentence} proposed a sentence-level RM, which segmented responses into individual sentences and computed rewards based on the difference between the outputs at each sentence’s start and end positions. To produce a final response-level reward, they introduced an attention mechanism that aggregated the sentence-level rewards.
Similarly, \citet{wang2024interpretable} presented a two-stage interpretable reward modeling framework. They first trained a multi-aspect ArmoRM on human-aligned dimensions such as honesty, verbosity, and safety. Then they employed a Mixture of Experts architecture with a gating mechanism to automatically select the most suitable reward dimension based on input context, which significantly improved model stability and alignment fidelity. \citet{dubois2023rewarded} proposed Rewarded Soup, a multi-policy strategy that trained separate models on different proxy rewards and linearly interpolated their weights to achieve Pareto-optimal generalization across the space of human preferences.

In summary, the design of fusion mechanisms evolved from static combinations to more adaptive, modular, and context-sensitive architectures, highlighting an ongoing effort to balance reliability, generalizability, and controllability in the alignment of LLMs.

\section{Reward model: functional roles under different optimization paradigms}
\label{5}
This section examines the distinct functional roles of reward models RMs in different LLM alignment paradigms, categorizing them into explicit and implicit forms. In RLHF, explicit RMs are trained on preference triplets to score responses and guide policy optimization via RL algorithms such as PPO, making them central to the optimization process. In \textbf{In-Context Learning} (ICL), explicit RMs function as external evaluators, assigning utility scores to candidate prompts or outputs for selection and filtering without updating model parameters, thus serving solely as decision aids. In \textbf{Direct Preference Optimization} (DPO) \cite{rafailov2023direct}, implicit RMs are embedded within the supervised optimization objective, integrating the reward signal directly into parameter updates without separate reward model training. Figure~\ref{fig3} illustrates this distinction, and Table \ref{table3} provides a comparative overview of how RMs are employed across these optimization paradigms.

\begin{figure*}[htbp]
	\centering
	\includegraphics[width=\textwidth]{F3.pdf}  % ???????????
	\caption{\label{fig3}Explicit reward modeling and implicit reward modeling represent two distinct approaches for optimizing LLM alignment, differing in their learning paradigms and implementation frameworks.}
\end{figure*}

\begin{table}[htbp]
	\caption{A comparison of LLM alignment methods from the perspective of optimization strategies.}
	\label{table3}
	\centering
	\footnotesize
	\begin{tabular}{
			>{\centering\arraybackslash}m{1.3cm}
			>{\raggedright\arraybackslash}m{3.2cm}
			>{\centering\arraybackslash}m{1.1cm}
			>{\centering\arraybackslash}m{1.5cm}
			>{\centering\arraybackslash}m{1.1cm}
			>{\raggedright\arraybackslash}m{3cm}
		}
		\toprule
		\textbf{Method} & \textbf{Input / Output} & \textbf{Uses RM?} & \textbf{Reward Category} & \textbf{Updates Parameters?} & \textbf{Note} \\
		\midrule
		SL (SFT) &
		Input: query \(x\)\newline
		Output: human-written label \(y^*\) &
		No &
		-- &
		Yes &
		Treats \(y^*\) as optimal; model learns to imitate human reference answers. \\
		
		SL (DPO) &
		Input: \(x\), pair of responses \((y^+, y^-)\)\newline
		Output: preference direction \(y^+ \succ y^-\) &
		Yes &
		Implicit RM &
		Yes &
		Trains model to prefer \(y^+\) over \(y^-\). \\
		
		RL &
		Input: \(x\)\newline
		Output: \(y \sim P_\theta(y \mid x)\), scored by RM \(R(x, y)\) &
		Yes &
		Explicit RM &
		Yes &
		Reward signal is used to train the model via policy optimization. \\
		
		ICL &
		Input: demo set \(C = \{(x_i, y_i)\}_{i=1}^k\), query \(x\)\newline
		Output: \(y \sim P_\theta(y \mid C, x)\) &
		Optional &
		Explicit RM (if used) &
		No &
		Reward signal guides the selection of demonstrations and evaluates or re-ranks candidate outputs. \\
		\bottomrule
	\end{tabular}
\end{table}

\subsection{Explicit reward modeling}
\subsubsection{RL-based LLM alignment frameworks}

RLHF typically involves three phases: (i) SFT, (ii) pre-training, and (iii) RM training and RL-based fine-tuning \cite{dong2024rlhf}. The RL fine-tuning phase then leverages the  feedback from the trained RM to optimize the pre-trained SFT model. The objective is to fine-tune the policy to maximize the expected reward under the learned RMs, while also incorporating a KL regularization term to ensure the updated policy remains close to the reference model. The overall objective can be expressed as:
\begin{equation}
\label{EQ8}
\max_{\pi_{\theta}}\mathbb{E}_{x\sim\mathcal{D},y\sim\pi_{\theta}(y|x)}[r_{\phi}(x,y)]-\beta\mathbb{D}_{\mathrm{KL}}[\pi_{\theta}(y\mid x)||\pi_{\mathrm{ref}}(y\mid x)]
\end{equation}

In this formation:

\begin{itemize}
	\item [\textbullet] 
	$\pi_\theta$ represents the policy of the language model, parameterized by $\theta$.
	\item [\textbullet]
	$\mathbb{E}_{x\sim\mathcal{D},y\sim\pi_\theta(y|x)}[r_\phi(x,y)]$ denotes the expected reward, where $r_\phi(x,y)$ is the RM that quantifies
	he alignment of the response $y$ with human preferences given the prompt $x$.
	\item [\textbullet]
	$\pi_\mathrm{ref}$ denotes the reference model, which serves as a baseline or starting point for the optimization.
	\item [\textbullet]
	$\mathbb{D}_\mathrm{KL}\left[\pi_\theta(y\mid x)\|\pi_\mathrm{ref}(y\mid x)\right]$ is the \textbf{Kullback-Leibler} (KL) divergence between the current policy $\pi_\theta$ and the
	reference model $\pi_{ref}.$
	\item [\textbullet]
	$\beta$ is a hyperparameter that balances the reward maximization with the KL divergence penalty.
	
\end{itemize}

While RLHF has achieved promising results in aligning LLMs with human preferences, PPO, its most commonly used optimization algorithm, remains limited by instability and sample inefficiency, particularly when compared SFT methods \cite{choshen2019weaknesses}. It is sensitive to hyperparameter configurations \cite{engstrom2020implementation}, susceptible to fluctuations in reward signals, and often fails to maintain consistent output characteristics such as response length. Furthermore, the reliance on an actor-critic architecture with a learnable value function increases the computational cost of training \cite{zhong2024dpo}. These limitations have hindered reproducibility, especially in open-source settings with constrained resources. In response, recent studies have proposed enhancements to PPO or explored alternative RL algorithms better suited for reward-driven fine-tuning.

\begin{itemize}
	\item [\textbullet] 
	Improving PPO:  \citet{zheng2023improving} clustered training samples into easy and hard categories using self-supervised similarity, and adjusted the KL regularization strength accordingly. This stratified reward shaping strategy enabled more stable and targeted optimization. CPPO \cite{zhang2024cppo} incorporated a reward-informed knowledge retention term into the PPO objective, which adjusted policy updates according to response novelty and prior alignment. This approach helped preserve desirable behaviors while encouraging adaptation to new reward signals. \citet{wu2023pairwise} proposed P3O to stabilize training by incorporating pairwise reward differences directly into the policy gradient computation, effectively reducing gradient variance arising from biased value estimation. \citet{santacroce2023efficient} addressed memory constraints in large-scale RLHF by sharing a common LLM backbone across reward, policy, value, and reference models, updating only lightweight adapters. This parameter-sharing strategy maintained RM fidelity while significantly improving scalability. \citet{shen2024contrastive} introduced contrastive rewards, an enhanced reward formulation that improves RLHF robustness by dynamically adjusting rewards based on sampled baselines and incorporating a penalty term for uncertain signals. This approach automatically adapted to task difficulty while mitigating noise in human feedback, leading to more stable policy optimization and better alignment performance compared to standard RMs.
	\citet{rafailov2023secrets} proposed PPO-Max, which enhances PPO's stability in RLHF by explicitly maximizing reward signals while maintaining training balance. The method demonstrated superior performance with sparse or noisy feedback.

	\item [\textbullet]

		Substituting PPO:	Rather than refining PPO, \citet{li2023remax} simplified the RLHF pipeline for auto-regressive language generation by removing non-essential components of PPO and reverted to a lightweight REINFORCE-style algorithm. They introduced ReMax, which leveraged entropy-regularized max-reward sampling to enhance training stability and sample efficiency, and outperformed PPO particularly in low-resource and off-policy regimes.  Similarly, REINFORCE-based methods alleviated the computational burden of critic networks by fully eliminating them. \citet{ahmadian2024back} proposed BackRLHF, which employed \textbf{REINFORCE Leave-One-Out} (RLOO) in conjunction with an auxiliary inverse dynamics model to infer user intent, thereby improving learning efficiency in resource-constrained settings through a lightweight and modular architecture. \citet{shao2024deepseekmath} designed \textbf{Group Relative Policy Optimization} (GRPO) to replace absolute reward signals with group-based relative feedback, which produced stronger gradient signals and improved stability in critic-free training. \citet{hu2025reinforcepp} proposed REINFORCE++, a critic-free RLHF algorithm that improved advantage estimation by using batch-normalized rewards as baselines. This design mitigated overfitting and reward hacking, and achieved robust generalization across RMs and chain-of-thought settings.

\end{itemize}

\subsection{ICL-based LLM alignment framework}

ICL is a paradigm that allows language models to learn tasks given only a few examples in the form of demonstration \cite{dong2023survey}. Formally, given a query input text \( x \) and a set of candidate answers \( Y = \{y_1, \ldots, y_m\} \), a pretrained language model takes the candidate answer with the maximum score as the prediction conditioned on a demonstration set \( C \). The set \( C \) contains an optional task instruction \( I \) and \( k \) demonstration examples, thus:
\begin{equation}
\label{EQ9}
C = \{I, s(x_1, y_1), \ldots, s(x_k, y_k)\}
\quad \text{or} \quad
C = \{s'(x_1, y_1, I), \ldots, s'(x_k, y_k, I)\},
\end{equation}
where \( s'(x_i, y_i, I) \) is an example written in natural language according to the task.

\begin{equation}
\label{EQ10}
\hat{y} =  \arg\max_{y_j \in Y} R(y_j, C, x)
\end{equation}
where \( \hat{y} \) denotes the final predicted label, i.e., the candidate answer with the highest probability, and \( R(y_j, C, x) \) is a scoring function that assesses the quality or relevance of each candidate \( y_j \) given the context \( C \) and the input query \( x \).

In addition to serving as a decision assessor for evaluating or re-ranking candidate outputs to improve alignment without parameter updates, RMs can also function as a utility function to guide prompt construction.

\subsubsection{Explicit RMs for prompt construction in ICL}
In ICL, prompt construction, including the selection, formatting, and ordering
of demonstration examples, is a critical determinant of downstream performance. While earlier approaches to prompt design typically depended on human heuristics or embedding-based similarity, the integration of RMs introduces a data-driven paradigm, enabling optimization of prompt inputs based on feedback signals aligned with task-specific objectives. In this context, the RM acts as a utility estimator, transforming prompt construction from a static template-design problem into a dynamic selection and synthesis process.

A prominent application of RMs lies in demonstration selection. For example, \citet{ye2023compositional} leveraged \textbf{Determinantal Point Processes} (DPPs) as a reward proxy to select diverse and representative in-context examples. \citet{du2024incontext} introduced an RL-driven context selection framework that treats prompt selection as a sequential decision-making task supervised by downstream reward signals. \citet{suo2024visual} used segmentation accuracy as feedback to guide stepwise demonstration selection, achieving better context-efficiency.

Beyond selection, RMs also aid in formatting and ordering demonstrations. \citet{zhang2024instruct} proposed reward-weighted ordering strategies in visual ICL by optimizing token-level segmentation accuracy. \citet{do2024prompt} introduced a generator-discriminator setup (adv-ICL) where the discriminator serves as an RM evaluating the correctness of model outputs elicited by generated prompts. \citet{zhou2023automatic} extended this idea in  \textbf{Automatic Prompt Engineer} (APE), where LLM-generated prompts are ranked using a task-specific RM to identify the most effective formulation. \citet{qian2024subsa} explored submodular optimization for annotation-efficient prompt construction, treating reward as a function of informativeness and diversity. In all these methods, RMs provide a way to systematically search for or synthesize high-performing prompts, converting the input space into an optimizable reward landscape.

\subsubsection{Explicit RMs for output evaluation in ICL}
Beyond guiding input construction, RMs play an increasingly important role in evaluating and aligning model outputs in ICL. Since ICL does not involve model parameter updates, post-generation supervision via RMs becomes essential for identifying the most reliable and aligned completions. In this phase, RMs act as task-specific critics that evaluate generated outputs for correctness, helpfulness, consistency, or other alignment objectives.

A fundamental use of RMs is in output re-ranking. \citet{li2025dora} introduced DORA, which used an RM to dynamically select completions that exhibit logical consistency and reasoning accuracy, even correcting high-confidence incorrect predictions. These examples illustrate how RMs can compensate for the lack of gradient-based learning in ICL by acting as post-hoc validators. RMs also serve as evaluation tools in benchmark construction. \citet{chen2024icleval} developed ICLEval, a benchmark that used RM-like metrics to evaluate ICL abilities in rule induction, copying, and memorization. \citet{yu2024rethinking} proposed a two-dimensional evaluation framework considering both performance via reward and configuration cost, offering a more holistic assessment of ICL alignment. \citet{honovich2022unnatural} used reward-driven evaluations to assess automatically generated task instructions, showing that RMs can serve as substitutes for human evaluation in instruction tuning. Recent works have further extended RMs to multi-pass reasoning and self-refinement. \citet{yuan2024self} used a reward heuristic to select the best answer among multiple iterative completions in Self-Refine, while \citet{lightman2023let} introduced stepwise rewards for intermediate reasoning steps to guide chain-of-thought alignment. 

Taken together, these studies illustrate the critical role that reward modeling plays in enhancing the robustness, transparency, and task alignment of outputs generated via ICL. By functioning as post-hoc evaluators, RMs provide a principled mechanism for output selection and validation, enabling improved performance without requiring parameter updates, thus offering a flexible and scalable alternative to traditional fully SFT approaches.

\subsection{Implicit reward modeling}
Explicit reward modeling in LLMs presents several inherent limitations, such as reward misspecification, high annotation costs, and optimization brittleness. As a result, an increasing number of researchers are exploring reward-free paradigms or alternative formulations that substitute or transform RMs into more efficient alignment objectives.

To reduce the complexity and cost of alignment, some researchers have focused on the initial phase of RLHF, SFT, and proposed a range of sophisticated SFT variations aiming to achieve comparable performance to RLHF  \cite{sun2023selfalign,wang2023selfinstruct,zhou2023lima}. Omitting $x$ for brevity, a general form of SFT alignment can be expressed as:
\begin{equation}
\arg\min_{\theta} - \mathbb{E}{p(y_w, y_l)} \left[ \log \pi{\theta}(y_w) - \log \pi_{\theta}(y_l) \right]
\propto \mathrm{KL}\left[ p(y_w) | \pi_{\theta}(y_w) \right] - \mathrm{KL}\left[ p(y_l) | \pi_{\theta}(y_l) \right],
\label{EQ11}
\end{equation}
where $y_w$ and $y_l$ denote the preferred (winning) and dispreferred (losing) responses respectively, sampled from the empirical joint preference distribution $p(y_w, y_l)$. $\pi_{\theta}(y)$ represents the model's policy parameterized by $\theta$, which assigns a probability to each response $y$. As a reward-free paradigm, it does not involve explicit reward modeling but directly learns to imitate preferred behaviors while unlearning dispreferred ones \cite{xu2024essence}. Hence, further elaboration is omitted. It is worth noting that while reward design offers solutions to many alignment challenges, it is not indispensable for achieving effective alignment, with reward-free methods increasingly emerging as a key research trend in LLM alignment.

Recent advances such as DPO \cite{rafailov2023direct} and implicit reward modeling illustrate the trend toward integrating preference learning more tightly into the optimization objective, eliminating the need for explicit reward computation while still guiding model behavior effectively. DPO offers a streamlined alternative to RLHF by eliminating the need for explicit reward modeling. Instead of learning a separate RM, DPO implicitly encodes human preferences directly within its objective function. Concretely, DPO uses the log-ratio of policy likelihoods, comparing preferred and dispreferred responses relative to a reference policy, as a proxy for reward, thereby reparameterizing the optimization target in terms of preference. This formulation not only simplifies the alignment pipeline but also improves training stability and efficiency.

Deriving from the KL-constrained reward maximization objective in Equation~\ref{EQ8}, DPO reformulates the optimization target as follows \cite{xiao2024comprehensive}:

\begin{equation}
\label{EQ12}
\pi_r(y\mid x) = \frac{1}{Z(x)} \,\pi_{\mathrm{ref}}(y\mid x)\,\exp\!\Bigl(\tfrac{1}{\beta}\,r(x,y)\Bigr)
\end{equation}

In this case,  
$
Z(x) \;=\;\sum_y \pi_{\mathrm{ref}}(y\mid x)\,\exp\!\Bigl(\tfrac{1}{\beta}\,r(x,y)\Bigr)
$
acts as the partition function that normalizes the policy distribution \(\pi_r(y\mid x)\). It is computed by summing the exponentiated reward scores weighted by the reference model’s distribution across every possible response \(y\).

Such summation guarantees that \(\pi_r(y\mid x)\) is a valid probability distribution; however, enumerating all candidate responses for each input \(x\) is impractical, requiring substantial computational resources when the response space is large. In practice, the DPO algorithm mitigates the need for explicit reward-model training and sampling from the language-model policy. By optimizing preferences directly, without the intermediate step of training an RM, DPO simplifies the training loop and reduces computational burden, offering a more efficient approach to language model alignment.

More precisely, the objective in Equation~\ref{EQ6} can be rearranged to isolate \(r(x,y)\), yielding:

\begin{equation}
\label{EQ13}
r(x,y) = \beta \,\log\frac{\pi_r(y\mid x)}{\pi_{\mathrm{ref}}(y\mid x)} \;+\;\beta \,\log Z(x)
\end{equation}
This reparameterization expresses the reward in terms of the optimized policy \(\pi_r\), the reference policy \(\pi_{\mathrm{ref}}\), and the normalization constant \(Z(x)\). Substituting into the original reward function \(r^*\) gives:

\begin{equation}
\label{EQ14}
r^*(x,y) = \beta \,\log\frac{\pi_r^*(y\mid x)}{\pi_{\mathrm{ref}}(y\mid x)} \;+\;\beta \,\log Z(x)
\end{equation}

Subsequently, the original loss of \(r^*\) is substituted with this reparameterization from Equation~\ref{EQ14}. Since the Bradley–Terry preference model depends only on reward differences between two completions, the human preference probability becomes:

\begin{equation}
\label{EQ15}
p^*\bigl(y_1 \succ y_2 \mid x\bigr)
= \frac{1}{\,1 + \exp\!\Bigl(\beta \log \tfrac{\pi^*(y_2\mid x)}{\pi_{\mathrm{ref}}(y_2\mid x)}
	- \beta \log \tfrac{\pi^*(y_1\mid x)}{\pi_{\mathrm{ref}}(y_1\mid x)}\Bigr)\,}.
\end{equation}

This simplifies via the sigmoid function as  $p^*(y_1 \succ y_2 \mid x) = \sigma\bigl(r^*(x,y_1) - r^*(x,y_2)\bigr),$ 
since \(Z(x)\) cancels out in the subtraction, resulting in a direct reward difference. After this simplification, a maximum likelihood objective for DPO can be formulated. The resulting policy loss for DPO is:
\begin{equation}
\label{EQ16}
\begin{split}
\mathcal{L}_{\mathrm{DPO}}(\pi_\theta;\pi_{\mathrm{ref}})
={} & -\mathbb{E}_{(x,y_w,y_l)\sim\mathcal{D}}
\Bigl[\log \sigma\bigl(\,
\beta \log \tfrac{\pi_\theta(y_w\mid x)}{\pi_{\mathrm{ref}}(y_w\mid x)}
- \beta \log \tfrac{\pi_\theta(y_l\mid x)}{\pi_{\mathrm{ref}}(y_l\mid x)}
\bigr)\Bigr].
\end{split}
\end{equation}

In this formulation:
\begin{itemize}
	\item \(\pi_\theta\) denotes the trainable policy of the target language model,
	\item \(\pi_{\mathrm{ref}}\) is the fixed reference policy,
	\item \(y_w\) and \(y_l\) refer to preferred and dispreferred responses, respectively.
\end{itemize}
This design enables DPO to model preferences directly through policy probabilities, without relying on explicit scalar rewards. As a result, alignment effectiveness is preserved while the training process is streamlined. 

Building upon this foundation, subsequent works have sought to refine DPO's objective function to strengthen the alignment signal. These improvements aim to enhance the granularity and expressiveness of the implicit reward, promote generalization, mitigate undesired behaviors such as reward hacking, and better reflect the structure of human preferences. \textbf{$\alpha$-DPO}~\cite{wu2024alphadpo} employed an adaptive preference distribution to balance the policy model and the reference model, thereby achieving personalized reward margins. Experimental results demonstrated its effectiveness as a surrogate optimization objective and its capability to balance alignment and generation diversity through KL divergence control. \textbf{$\beta$-DPO}~\cite{wu2024beta}, dynamically adjusted the trade-off parameter  $\beta$ at the batch level based on data informativeness, and employs  $\beta$‑guided data filtering to mitigate the influence of noisy or outlier preference pairs. \textbf{T-DPO}~\cite{zeng2024token} incorporated forward KL divergence constraints for each token, improving alignment and diversity.Bradley-Terry model was utilized for a token-based reward system to preserve simplicity without the need for explicit reward modeling. To address the lack of fine-grained process supervision in long-chain mathematical reasoning tasks, \textbf{Step-DPO}~\cite{lai2024stepdpo}  treated each individual reasoning step as an independent unit for preference optimization. By shifting the optimization granularity from holistic answer-level evaluation to step-wise supervision, Step-DPO enabled the model to accurately identify subtle errors within complex reasoning chains, thereby enhancing its ability to align with human expectations in multi-step problem-solving scenarios.

For a comprehensive overview of DPO and its variants to viewed through the evolution of reward modeling, interested readers may refer to \cite{xu2024dposurvey}. Taken together, these advances demonstrate the potential of optimizing objective functions to model reward signals implicitly. Through richer representations of preference strength, fine-grained token-level comparisons, robustness to pathological behavior, and improved optimization strategies, DPO and its variants offer a lightweight yet effective approach to preference-based alignment without relying on explicitly learned RMs.

\section{Discussion}
\label{6}
Table~\ref{table 1} presents a systematic classification of representative LLM alignment studies, structured across three key dimensions: (i) feedback sources, (ii) optimization paradigms, and (iii) RM characteristics.
Based on this classification, we observe several significant trends in the development of reward modeling techniques.

\subsection{Progress of reward model}

The analysis highlights several notable trends in the evolution of reward design. Initially, reward mechanisms were rule-based, manually encoding rewards based on predefined criteria. Over time, reward modeling shifted toward data-driven approaches, learning reward signals directly from behavioral or annotated data. This transition also introduced richer, multidimensional feedback expressed in natural language, moving beyond simple scalar rewards. These richer signals allow models to better capture nuanced human preferences and foster more effective alignment. In parallel, reward functions have evolved from explicit formulations to more flexible, implicit mechanisms that adapt dynamically to diverse tasks without rigid predefinitions. Additionally, the granularity of RMs has evolved. Early reward functions were general and coarse, but now fine-grained reward structures are increasingly used, providing detailed feedback at various interaction levels, such as token-level, multi-objective rewards and hybrid. This evolution enables models to be more sensitive to human feedback specifics and enhances their adaptability to complex tasks. 

Advancements in reward design have thereby driven two prominent trends in LLM alignment. First, there has been a notable shift from RL-based optimization (e.g., PPO-based RLHF) to RL-free alternatives, such as SL-based (e.g., DPO) and ICL-based alignment. These methods offer greater stability, efficiency, and scalability, especially in scenarios with limited or noisy feedback. Second, the improved flexibility and expressiveness of modern reward mechanisms have significantly enhanced the capacity of alignment frameworks to handle multi-modal and multi-task settings, where static, scalar, and task-specific rewards often fall short. Collectively, these shifts mark a fundamental transformation in aligning LLMs with human values under complex, real-world conditions.

\subsection{Future directions of reward model}
Looking beyond current trends, reward design for LLM alignment is expected to move from static rule-following to dynamic value co-creation. Future approaches will focus on learning values from collective human interactions, such as dialogues, feedback, and community input, rather than relying solely on static annotations. This shift calls for context-aware and temporally adaptive RMs that reflect evolving, diverse, and sometimes conflicting human values.
Additionally, socially interactive and meta-reward learning frameworks may enable LLMs to continually update alignment objectives based on real-time feedback and cultural context. Rather than serving only as tools for task completion, reward mechanisms will increasingly support ethical reasoning and value negotiation. In high-stakes or multi-agent scenarios, RMs must also incorporate governance norms and collective decision-making principles.
Ultimately, future reward design will play a central role in building AI systems that align not just with static preferences, but with the dynamic values of human society.

\section{Conclusions}
\label{7}
The challenges faced by LLMs and their alignment processes, including issues arising from the introduction of reward mechanisms, can be addressed through well-designed reward modeling. A review of recent studies indicates that the evolution of LLM alignment can be regarded as a continuous process of exploring, evaluating, and refining reward design strategies. As alignment scenarios grow more complex, spanning multi-objective, multi-task, and multi-modal contexts, the role of reward models will increasingly shift toward being adaptive, context-aware systems that bridge dynamic human values and autonomous model behaviors.

% \bmhead{Acknowledgements}

% Acknowledgements are not compulsory. Where included they should be brief. Grant or contribution numbers may be acknowledged.

% \section*{Declarations}

% \begin{itemize}

% \item Author contributions

% The completion of this paper was a result of the collaborative eforts of all. Miaomiao: Conceptualization, Visualization, Writing-Original Draft. Yanqiu: Conceptualization, Supervision, Writing-Review \& Editing. Zhibin: Supervision, Writing-Review \& Editing. Usman: Conceptualization, Supervision, Project Administration, Funding Acquisition, Writing-Review \& Editing, 
% All authors reviewed and approved the final version of the manuscript.

% xxx
% x
% x
% x
% x

% No datasets were generated or analysed during the current study.

% \item Ethics approval and consent to participate

% Not applicable.
% \item Consent for publication

% Not applicable.
% \item Competing interests

% The authors declare no competing interests.
% \end{itemize}

\begin{appendices}

\begin{sidewaystable}
	\centering
	\caption{Systematic analysis of LLM alignment studies: feedback source, optimization, learning paradigm, and RM architectures (Alphabetical by Author)}
	\label{table 1}
	\footnotesize
	\setlength{\tabcolsep}{1pt}
	\begin{tabular*}{\textwidth}{@{\extracolsep{\fill}}l*{11}{c}@{}}
		\toprule
		\multirow{2}{*}{Paper} & 
		\multicolumn{2}{c}{Feedback} & 
		\multicolumn{3}{c}{Optimization} & 
		\multicolumn{6}{c}{Reward Design} \\
		\cmidrule(lr){2-3} \cmidrule(lr){4-6} \cmidrule(lr){7-12}
		& \shortstack{Human feedback}  & \shortstack{AI feedback} & \shortstack{SL} & \shortstack{ RL} & \shortstack{ICL} & \shortstack{Rule-based \\RM} & \shortstack{Data-driven \\RM} & \shortstack{Numerical \\RM} & \shortstack{Non-Numerical \\RM} & \shortstack{Explicit \\RM} & \shortstack{Implicit\\RM} \\
		\midrule
		\citet{ahmadian2024back} & $\checkmark$ & & & $\checkmark$ & & & $\checkmark$ & $\checkmark$ & & $\checkmark$ & \\
		\citet{akyurek2023rl4f} & & $\checkmark$ & & $\checkmark$ & & & $\checkmark$ & & $\checkmark$ & $\checkmark$ & \\
		\citet{azar2024general} & $\checkmark$ & & & $\checkmark$ & & & $\checkmark$ & $\checkmark$ & & $\checkmark$ & \\
		\citet{bai2022training} & $\checkmark$ & & & $\checkmark$ & & & $\checkmark$ & $\checkmark$ & & $\checkmark$ & \\
		\citet{cao2024beyond} & & $\checkmark$ & & $\checkmark$ & & & $\checkmark$ & $\checkmark$ & & $\checkmark$ & \\
		\citet{chan2024dense} & $\checkmark$ & & & $\checkmark$ & & & $\checkmark$ & $\checkmark$ & & $\checkmark$ & \\
		\citet{cheng2023adversarial} & $\checkmark$ & & & $\checkmark$ & & & $\checkmark$ & $\checkmark$ & & $\checkmark$ & \\
		\citet{coste2023reward} & $\checkmark$ & & & $\checkmark$ & & & $\checkmark$ & $\checkmark$ & & $\checkmark$ & \\
		\citet{cui2023ultrafeedback} & & $\checkmark$ & & $\checkmark$ & & & $\checkmark$ & & $\checkmark$ & $\checkmark$ & \\
		\citet{dai2023safe} & $\checkmark$ & & & $\checkmark$ & & & $\checkmark$ & $\checkmark$ & & $\checkmark$ & \\
		\citet{dong2023steerlm} & $\checkmark$ & & $\checkmark$ & & & & & & & & $\checkmark$ \\
		\citet{ethayarajh2024kto} & $\checkmark$ & & $\checkmark$ & & & & & & & & $\checkmark$ \\
		\citet{hong2024orpo} & $\checkmark$ & & $\checkmark$ & & & & & & & & $\checkmark$ \\
		\citet{kim2023aligning} & $\checkmark$ & & & $\checkmark$ & & & $\checkmark$ & $\checkmark$ & & $\checkmark$ & \\
		\citet{lee2023rlaif} & & $\checkmark$ & & $\checkmark$ & & & $\checkmark$ & $\checkmark$ & & $\checkmark$ & \\
		\citet{li2023remax} & $\checkmark$ & & & $\checkmark$ & & & $\checkmark$ & $\checkmark$ & & $\checkmark$ & \\
		\citet{mahan2024generative} & & $\checkmark$ & $\checkmark$ & & & & $\checkmark$ & $\checkmark$ & & $\checkmark$ & \\
		\citet{moskovitz2023confronting} & $\checkmark$ & & & $\checkmark$ & & & $\checkmark$ & $\checkmark$ & & $\checkmark$ & \\
		\citet{mu2024rule} & & $\checkmark$ & & $\checkmark$ & & $\checkmark$ & & $\checkmark$ & $\checkmark$ & $\checkmark$ & \\
		\citet{pang2404iterative} & $\checkmark$ & & $\checkmark$ & & & & & & & & $\checkmark$ \\
		\citet{park2024disentangling} & $\checkmark$ & & $\checkmark$ & & & & & & & & $\checkmark$ \\
		\citet{rafailov2024direct} & $\checkmark$ & & $\checkmark$ & & & & & & & & $\checkmark$ \\
		\citet{rame2023rewarded} & $\checkmark$ & & & $\checkmark$ & & & $\checkmark$ & $\checkmark$ & & $\checkmark$ & \\
		\citet{santacroce2023efficient} & $\checkmark$ & & & $\checkmark$ & & & $\checkmark$ & $\checkmark$ & & $\checkmark$ & \\
		\citet{scheid2024optimal} & $\checkmark$ & & $\checkmark$ & & & & $\checkmark$ & $\checkmark$ & & $\checkmark$ & \\
		\citet{wu2023fine} & $\checkmark$ & & & $\checkmark$ & & & $\checkmark$ & $\checkmark$ & & $\checkmark$ & \\
		\citet{xu2024dpo} & $\checkmark$ & & $\checkmark$ & & & & & & & & $\checkmark$ \\
		\citet{ye2023compositional} & & & & & $\checkmark$ & & & & & $\checkmark$ & \\
		\citet{yang2024selective} & $\checkmark$ & & $\checkmark$ & & & & $\checkmark$ & $\checkmark$ & & $\checkmark$ & \\
		\citet{yin2025constrain} & $\checkmark$ & & $\checkmark$ & & & & & & & & $\checkmark$ \\
		\citet{yoon2024tlcr} & $\checkmark$ & & & $\checkmark$ & & & $\checkmark$ & $\checkmark$ & & $\checkmark$ & \\
		\citet{zeng2024token} & $\checkmark$ & & $\checkmark$ & & & & $\checkmark$ & $\checkmark$ & & $\checkmark$ & \\
		\citet{zhang2024instruct} & & & & & $\checkmark$ & & & & & $\checkmark$ & \\
		\citet{zhang2024generative} & $\checkmark$ & & $\checkmark$ & & & & $\checkmark$ & $\checkmark$ & & $\checkmark$ & \\
		\bottomrule
	\end{tabular*}
\end{sidewaystable}




\end{appendices}

\clearpage
\bibliography{filtered_reference.bib}% common bib file


\begin{thebibliography}{130}
\ifx \bisbn   \undefined \def \bisbn  #1{ISBN #1}\fi
\ifx \binits  \undefined \def \binits#1{#1}\fi
\ifx \bauthor  \undefined \def \bauthor#1{#1}\fi
\ifx \batitle  \undefined \def \batitle#1{#1}\fi
\ifx \bjtitle  \undefined \def \bjtitle#1{#1}\fi
\ifx \bvolume  \undefined \def \bvolume#1{\textbf{#1}}\fi
\ifx \byear  \undefined \def \byear#1{#1}\fi
\ifx \bissue  \undefined \def \bissue#1{#1}\fi
\ifx \bfpage  \undefined \def \bfpage#1{#1}\fi
\ifx \blpage  \undefined \def \blpage #1{#1}\fi
\ifx \burl  \undefined \def \burl#1{\textsf{#1}}\fi
\ifx \doiurl  \undefined \def \doiurl#1{\url{https://doi.org/#1}}\fi
\ifx \betal  \undefined \def \betal{\textit{et al.}}\fi
\ifx \binstitute  \undefined \def \binstitute#1{#1}\fi
\ifx \binstitutionaled  \undefined \def \binstitutionaled#1{#1}\fi
\ifx \bctitle  \undefined \def \bctitle#1{#1}\fi
\ifx \beditor  \undefined \def \beditor#1{#1}\fi
\ifx \bpublisher  \undefined \def \bpublisher#1{#1}\fi
\ifx \bbtitle  \undefined \def \bbtitle#1{#1}\fi
\ifx \bedition  \undefined \def \bedition#1{#1}\fi
\ifx \bseriesno  \undefined \def \bseriesno#1{#1}\fi
\ifx \blocation  \undefined \def \blocation#1{#1}\fi
\ifx \bsertitle  \undefined \def \bsertitle#1{#1}\fi
\ifx \bsnm \undefined \def \bsnm#1{#1}\fi
\ifx \bsuffix \undefined \def \bsuffix#1{#1}\fi
\ifx \bparticle \undefined \def \bparticle#1{#1}\fi
\ifx \barticle \undefined \def \barticle#1{#1}\fi
\bibcommenthead
\ifx \bconfdate \undefined \def \bconfdate #1{#1}\fi
\ifx \botherref \undefined \def \botherref #1{#1}\fi
\ifx \url \undefined \def \url#1{\textsf{#1}}\fi
\ifx \bchapter \undefined \def \bchapter#1{#1}\fi
\ifx \bbook \undefined \def \bbook#1{#1}\fi
\ifx \bcomment \undefined \def \bcomment#1{#1}\fi
\ifx \oauthor \undefined \def \oauthor#1{#1}\fi
\ifx \citeauthoryear \undefined \def \citeauthoryear#1{#1}\fi
\ifx \endbibitem  \undefined \def \endbibitem {}\fi
\ifx \bconflocation  \undefined \def \bconflocation#1{#1}\fi
\ifx \arxivurl  \undefined \def \arxivurl#1{\textsf{#1}}\fi
\csname PreBibitemsHook\endcsname

%%% 1
\bibitem[\protect\citeauthoryear{OpenAI}{2023}]{openai2023gpt4}
\begin{botherref}
\oauthor{\bsnm{OpenAI}}:
Gpt-4 technical report.
arXiv preprint arXiv:2303.08774
(2023)
\end{botherref}
\endbibitem

%%% 2
\bibitem[\protect\citeauthoryear{Anthropic}{2023}]{anthropic2023claude}
\begin{botherref}
\oauthor{\bsnm{Anthropic}}:
Introducing Claude.
https://www.anthropic.com/index/introducing-claude
(2023)
\end{botherref}
\endbibitem

%%% 3
\bibitem[\protect\citeauthoryear{DeepMind}{2023}]{google2023gemini}
\begin{botherref}
\oauthor{\bsnm{DeepMind}, \binits{G.}}:
Google Gemini: Our largest and most capable AI models.
https://deepmind.google/technologies/gemini/
(2023)
\end{botherref}
\endbibitem

%%% 4
\bibitem[\protect\citeauthoryear{Vaswani et~al.}{2017}]{vaswani2017attention}
\begin{bchapter}
\bauthor{\bsnm{Vaswani}, \binits{A.}},
\bauthor{\bsnm{Shazeer}, \binits{N.}},
\bauthor{\bsnm{Parmar}, \binits{N.}},
\bauthor{\bsnm{Uszkoreit}, \binits{J.}},
\bauthor{\bsnm{Jones}, \binits{L.}},
\bauthor{\bsnm{Gomez}, \binits{A.N.}}, \betal:
\bctitle{Attention is all you need}.
In: \bbtitle{Advances in Neural Information Processing Systems},
vol. \bseriesno{30},
pp. \bfpage{5998}--\blpage{6008}
(\byear{2017})
\end{bchapter}
\endbibitem

%%% 5
\bibitem[\protect\citeauthoryear{Bommasani et~al.}{2021}]{bommasani2021opportunities}
\begin{botherref}
\oauthor{\bsnm{Bommasani}, \binits{R.}},
\oauthor{\bsnm{Hudson}, \binits{D.A.}},
\oauthor{\bsnm{Adeli}, \binits{E.}},
\oauthor{\bsnm{Altman}, \binits{R.B.}},
\oauthor{\bsnm{Arora}, \binits{S.}},
\oauthor{\bsnm{Arx}, \binits{S.}}, et al.:
On the opportunities and risks of foundation models.
arXiv preprint arXiv:2108.07258
(2021)
\end{botherref}
\endbibitem

%%% 6
\bibitem[\protect\citeauthoryear{Wang et~al.}{2024}]{wang2024large}
\begin{botherref}
\oauthor{\bsnm{Wang}, \binits{S.}},
\oauthor{\bsnm{Xu}, \binits{T.}},
\oauthor{\bsnm{Li}, \binits{H.}},
\oauthor{\bsnm{Zhang}, \binits{C.}},
\oauthor{\bsnm{Liang}, \binits{J.}},
\oauthor{\bsnm{Tang}, \binits{J.}},
\oauthor{\bsnm{Yu}, \binits{P.S.}},
\oauthor{\bsnm{Wen}, \binits{Q.}}:
Large language models for education: A survey and outlook.
arXiv preprint arXiv:2403.18105
(2024)
\end{botherref}
\endbibitem

%%% 7
\bibitem[\protect\citeauthoryear{Askell et~al.}{2021}]{askell2021general}
\begin{botherref}
\oauthor{\bsnm{Askell}, \binits{A.}},
\oauthor{\bsnm{Bai}, \binits{Y.}},
\oauthor{\bsnm{Chen}, \binits{A.}},
\oauthor{\bsnm{Drain}, \binits{D.}},
\oauthor{\bsnm{Ganguli}, \binits{D.}},
\oauthor{\bsnm{Hernandez}, \binits{E.}},
\oauthor{\bsnm{Kaplan}, \binits{J.}},
\oauthor{\bsnm{Henighan}, \binits{T.}},
\oauthor{\bsnm{Legg}, \binits{S.}},
\oauthor{\bsnm{Milani}, \binits{S.}}, et al.:
A general language assistant as a laboratory for alignment.
arXiv preprint arXiv:2112.00861
(2021)
\end{botherref}
\endbibitem

%%% 8
\bibitem[\protect\citeauthoryear{Ji et~al.}{2023}]{ji2023survey}
\begin{barticle}
\bauthor{\bsnm{Ji}, \binits{Z.}},
\bauthor{\bsnm{Lee}, \binits{N.}},
\bauthor{\bsnm{Frieske}, \binits{R.}},
\bauthor{\bsnm{Yu}, \binits{T.}},
\bauthor{\bsnm{Su}, \binits{D.}},
\bauthor{\bsnm{Xu}, \binits{Y.}}, \betal:
\batitle{Survey of hallucination in natural language generation}.
\bjtitle{ACM Computing Surveys}
\bvolume{55}(\bissue{12}),
\bfpage{1}--\blpage{38}
(\byear{2023})
\end{barticle}
\endbibitem

%%% 9
\bibitem[\protect\citeauthoryear{Gehman et~al.}{2020}]{gehman2020realtoxicityprompts}
\begin{bchapter}
\bauthor{\bsnm{Gehman}, \binits{S.}},
\bauthor{\bsnm{Gururangan}, \binits{S.}},
\bauthor{\bsnm{Sap}, \binits{M.}},
\bauthor{\bsnm{Choi}, \binits{Y.}},
\bauthor{\bsnm{Smith}, \binits{N.A.}}:
\bctitle{Realtoxicityprompts: Evaluating neural toxic degeneration in language models}.
In: \bbtitle{Findings of the Association for Computational Linguistics: EMNLP 2020},
pp. \bfpage{3356}--\blpage{3369}
(\byear{2020}).
\burl{https://aclanthology.org/2020.findings-emnlp.301/}
\end{bchapter}
\endbibitem

%%% 10
\bibitem[\protect\citeauthoryear{Bender et~al.}{2021}]{bender2021dangers}
\begin{botherref}
\oauthor{\bsnm{Bender}, \binits{E.M.}},
\oauthor{\bsnm{Gebru}, \binits{T.}},
\oauthor{\bsnm{McMillan-Major}, \binits{A.}},
\oauthor{\bsnm{Shmitchell}, \binits{S.}}:
On the dangers of stochastic parrots: Can language models be too big?
Proceedings of the 2021 ACM Conference on Fairness, Accountability, and Transparency (FAccT),
610--623
(2021)
\end{botherref}
\endbibitem

%%% 11
\bibitem[\protect\citeauthoryear{Gabriel}{2020}]{gabriel2020artificial}
\begin{barticle}
\bauthor{\bsnm{Gabriel}, \binits{I.}}:
\batitle{Artificial intelligence, values, and alignment}.
\bjtitle{Minds and Machines}
\bvolume{30}(\bissue{3}),
\bfpage{411}--\blpage{437}
(\byear{2020})
\end{barticle}
\endbibitem

%%% 12
\bibitem[\protect\citeauthoryear{Carlini et~al.}{2023}]{carlini2023extracting}
\begin{bchapter}
\bauthor{\bsnm{Carlini}, \binits{N.}},
\bauthor{\bsnm{Hayes}, \binits{J.}},
\bauthor{\bsnm{Nasr}, \binits{M.}},
\bauthor{\bsnm{Jagielski}, \binits{M.}},
\bauthor{\bsnm{Sehwag}, \binits{V.}},
\bauthor{\bsnm{Tram{\`e}r}, \binits{F.}}, \betal:
\bctitle{Extracting training data from diffusion models}.
In: \bbtitle{Proceedings of the 32nd USENIX Security Symposium},
pp. \bfpage{5253}--\blpage{5270}
(\byear{2023}).
\burl{https://arxiv.org/abs/2301.13188}
\end{bchapter}
\endbibitem

%%% 13
\bibitem[\protect\citeauthoryear{Christiano et~al.}{2017}]{christiano2017deep}
\begin{botherref}
\oauthor{\bsnm{Christiano}, \binits{P.F.}},
\oauthor{\bsnm{Leike}, \binits{J.}},
\oauthor{\bsnm{Brown}, \binits{T.}},
\oauthor{\bsnm{Martic}, \binits{M.}},
\oauthor{\bsnm{Legg}, \binits{S.}},
\oauthor{\bsnm{Amodei}, \binits{D.}}:
Deep reinforcement learning from human preferences.
Advances in neural information processing systems
\textbf{30}
(2017)
\end{botherref}
\endbibitem

%%% 14
\bibitem[\protect\citeauthoryear{Ouyang et~al.}{2022}]{ouyang2022training}
\begin{barticle}
\bauthor{\bsnm{Ouyang}, \binits{L.}},
\bauthor{\bsnm{Wu}, \binits{J.}},
\bauthor{\bsnm{Jiang}, \binits{X.}},
\bauthor{\bsnm{Almeida}, \binits{D.}},
\bauthor{\bsnm{Wainwright}, \binits{C.}},
\bauthor{\bsnm{Mishkin}, \binits{P.}},
\bauthor{\bsnm{Zhang}, \binits{C.}},
\bauthor{\bsnm{Agarwal}, \binits{S.}},
\bauthor{\bsnm{Slama}, \binits{A.}},
\bauthor{\bsnm{Ray}, \binits{C.}}, \betal:
\batitle{Training language models to follow instructions with human feedback}.
\bjtitle{Advances in neural information processing systems}
\bvolume{35},
\bfpage{27730}--\blpage{27744}
(\byear{2022})
\end{barticle}
\endbibitem

%%% 15
\bibitem[\protect\citeauthoryear{Schulman et~al.}{2017}]{schulman2017proximal}
\begin{botherref}
\oauthor{\bsnm{Schulman}, \binits{J.}},
\oauthor{\bsnm{Wolski}, \binits{F.}},
\oauthor{\bsnm{Dhariwal}, \binits{P.}},
\oauthor{\bsnm{Radford}, \binits{A.}},
\oauthor{\bsnm{Klimov}, \binits{O.}}:
Proximal policy optimization algorithms.
arXiv preprint arXiv:1707.06347
(2017)
\end{botherref}
\endbibitem

%%% 16
\bibitem[\protect\citeauthoryear{Ziegler et~al.}{2019}]{ziegler2019fine}
\begin{botherref}
\oauthor{\bsnm{Ziegler}, \binits{D.M.}},
\oauthor{\bsnm{Stiennon}, \binits{N.}},
\oauthor{\bsnm{Wu}, \binits{J.}},
\oauthor{\bsnm{Brown}, \binits{T.B.}},
\oauthor{\bsnm{Radford}, \binits{A.}},
\oauthor{\bsnm{Amodei}, \binits{D.}}:
Fine-tuning language models from human preferences.
arXiv preprint arXiv:1909.08593
(2019)
\end{botherref}
\endbibitem

%%% 17
\bibitem[\protect\citeauthoryear{Jaques et~al.}{2019}]{jaques2019way}
\begin{botherref}
\oauthor{\bsnm{Jaques}, \binits{N.}},
\oauthor{\bsnm{Ghandeharioun}, \binits{A.}},
\oauthor{\bsnm{Shen}, \binits{J.H.}},
\oauthor{\bsnm{Ferguson}, \binits{C.}},
\oauthor{\bsnm{Lapedriza}, \binits{A.}},
\oauthor{\bsnm{Jones}, \binits{N.}},
\oauthor{\bsnm{Gu}, \binits{S.}},
\oauthor{\bsnm{Picard}, \binits{R.}}:
Way off-policy batch deep reinforcement learning of implicit human preferences in dialog.
arXiv preprint arXiv:1907.00456
(2019)
\end{botherref}
\endbibitem

%%% 18
\bibitem[\protect\citeauthoryear{Amodei et~al.}{2016}]{amodei2016concrete}
\begin{botherref}
\oauthor{\bsnm{Amodei}, \binits{D.}},
\oauthor{\bsnm{Olah}, \binits{C.}},
\oauthor{\bsnm{Steinhardt}, \binits{J.}},
\oauthor{\bsnm{Christiano}, \binits{P.}},
\oauthor{\bsnm{Schulman}, \binits{J.}},
\oauthor{\bsnm{Man{\'e}}, \binits{D.}}:
Concrete problems in ai safety.
arXiv preprint arXiv:1606.06565
(2016)
\end{botherref}
\endbibitem

%%% 19
\bibitem[\protect\citeauthoryear{Bai et~al.}{2022}]{bai2022training}
\begin{botherref}
\oauthor{\bsnm{Bai}, \binits{Y.}},
\oauthor{\bsnm{Jones}, \binits{A.}},
\oauthor{\bsnm{Ndousse}, \binits{K.}},
\oauthor{\bsnm{Askell}, \binits{A.}},
\oauthor{\bsnm{Chen}, \binits{A.}},
\oauthor{\bsnm{DasSarma}, \binits{N.}},
\oauthor{\bsnm{Drain}, \binits{D.}},
\oauthor{\bsnm{Fort}, \binits{S.}},
\oauthor{\bsnm{Ganguli}, \binits{D.}},
\oauthor{\bsnm{Henighan}, \binits{T.}}, et al.:
Training a helpful and harmless assistant with reinforcement learning from human feedback.
arXiv preprint arXiv:2204.05862
(2022)
\end{botherref}
\endbibitem

%%% 20
\bibitem[\protect\citeauthoryear{Zheng et~al.}{2023}]{zheng2023secrets}
\begin{botherref}
\oauthor{\bsnm{Zheng}, \binits{R.}},
\oauthor{\bsnm{Dou}, \binits{S.}},
\oauthor{\bsnm{Gao}, \binits{S.}},
\oauthor{\bsnm{Hua}, \binits{Y.}},
\oauthor{\bsnm{Shen}, \binits{W.}},
\oauthor{\bsnm{Wang}, \binits{B.}},
\oauthor{\bsnm{Liu}, \binits{Y.}},
\oauthor{\bsnm{Jin}, \binits{S.}},
\oauthor{\bsnm{Liu}, \binits{Q.}},
\oauthor{\bsnm{Zhou}, \binits{Y.}},
\oauthor{\bsnm{Xiong}, \binits{L.}},
\oauthor{\bsnm{Chen}, \binits{L.}},
\oauthor{\bsnm{Xi}, \binits{Z.}},
\oauthor{\bsnm{Xu}, \binits{N.}},
\oauthor{\bsnm{Lai}, \binits{W.}},
\oauthor{\bsnm{Zhu}, \binits{M.}},
\oauthor{\bsnm{Chang}, \binits{C.}},
\oauthor{\bsnm{Yin}, \binits{Z.}},
\oauthor{\bsnm{Weng}, \binits{R.}},
\oauthor{\bsnm{Cheng}, \binits{W.}},
\oauthor{\bsnm{Huang}, \binits{H.}},
\oauthor{\bsnm{Sun}, \binits{T.}},
\oauthor{\bsnm{Yan}, \binits{H.}},
\oauthor{\bsnm{Gui}, \binits{T.}},
\oauthor{\bsnm{Zhang}, \binits{Q.}},
\oauthor{\bsnm{Qiu}, \binits{X.}},
\oauthor{\bsnm{Huang}, \binits{X.}}:
Secrets of rlhf in large language models part i: Ppo.
arXiv preprint arXiv:2307.04964
(2023)
\end{botherref}
\endbibitem

%%% 21
\bibitem[\protect\citeauthoryear{Skalse et~al.}{2022}]{skalse2022defining}
\begin{bchapter}
\bauthor{\bsnm{Skalse}, \binits{J.}},
\bauthor{\bsnm{Howe}, \binits{N.H.R.}},
\bauthor{\bsnm{Krasheninnikov}, \binits{D.}},
\bauthor{\bsnm{Krueger}, \binits{D.}}:
\bctitle{Defining and characterizing reward hacking}.
In: \bbtitle{Advances in Neural Information Processing Systems},
vol. \bseriesno{35}
(\byear{2022}).
\burl{https://arxiv.org/abs/2209.13085}
\end{bchapter}
\endbibitem

%%% 22
\bibitem[\protect\citeauthoryear{Gao et~al.}{2023}]{gao2023scaling}
\begin{bchapter}
\bauthor{\bsnm{Gao}, \binits{L.}},
\bauthor{\bsnm{Schulman}, \binits{J.}},
\bauthor{\bsnm{Hilton}, \binits{J.}}:
\bctitle{Scaling laws for reward model overoptimization}.
In: \bbtitle{Proceedings of the 40th International Conference on Machine Learning}.
\bsertitle{Proceedings of Machine Learning Research},
vol. \bseriesno{202},
pp. \bfpage{10835}--\blpage{10866}
(\byear{2023}).
\burl{https://proceedings.mlr.press/v202/gao23h.html}
\end{bchapter}
\endbibitem

%%% 23
\bibitem[\protect\citeauthoryear{Tsimpoukelli et~al.}{2021}]{tsimpoukelli2021multimodal}
\begin{bchapter}
\bauthor{\bsnm{Tsimpoukelli}, \binits{M.}},
\bauthor{\bsnm{Menick}, \binits{J.}},
\bauthor{\bsnm{Cabi}, \binits{S.}},
\bauthor{\bsnm{Eslami}, \binits{S.M.A.}},
\bauthor{\bsnm{Vinyals}, \binits{O.}},
\bauthor{\bsnm{Hill}, \binits{F.}}:
\bctitle{Multimodal few-shot learning with frozen language models}.
In: \bbtitle{Advances in Neural Information Processing Systems},
vol. \bseriesno{34},
pp. \bfpage{200}--\blpage{212}
(\byear{2021}).
\burl{https://arxiv.org/abs/2106.13884}
\end{bchapter}
\endbibitem

%%% 24
\bibitem[\protect\citeauthoryear{Yao et~al.}{2023}]{yao2023react}
\begin{bchapter}
\bauthor{\bsnm{Yao}, \binits{S.}},
\bauthor{\bsnm{Zhao}, \binits{J.}},
\bauthor{\bsnm{Yu}, \binits{D.}},
\bauthor{\bsnm{Du}, \binits{N.}},
\bauthor{\bsnm{Shafran}, \binits{I.}},
\bauthor{\bsnm{Narasimhan}, \binits{K.}},
\bauthor{\bsnm{Cao}, \binits{Y.}}:
\bctitle{React: Synergizing reasoning and acting in language models}.
In: \bbtitle{Proceedings of the 11th International Conference on Learning Representations (ICLR)}
(\byear{2023}).
\burl{https://arxiv.org/abs/2210.03629}
\end{bchapter}
\endbibitem

%%% 25
\bibitem[\protect\citeauthoryear{Xu et~al.}{2023}]{xu2023multiinstruct}
\begin{bchapter}
\bauthor{\bsnm{Xu}, \binits{Z.}},
\bauthor{\bsnm{Shen}, \binits{Y.}},
\bauthor{\bsnm{Huang}, \binits{L.}}:
\bctitle{Multiinstruct: Improving multi-modal zero-shot learning via instruction tuning}.
In: \bbtitle{Proceedings of the 61st Annual Meeting of the Association for Computational Linguistics (Volume 1: Long Papers)},
pp. \bfpage{11445}--\blpage{11465}
(\byear{2023}).
\doiurl{10.18653/v1/2023.acl-long.641} .
\burl{https://aclanthology.org/2023.acl-long.641/}
\end{bchapter}
\endbibitem

%%% 26
\bibitem[\protect\citeauthoryear{Shen et~al.}{2023}]{shen2023large}
\begin{botherref}
\oauthor{\bsnm{Shen}, \binits{T.}},
\oauthor{\bsnm{Jin}, \binits{R.}},
\oauthor{\bsnm{Huang}, \binits{Y.}},
\oauthor{\bsnm{Liu}, \binits{C.}},
\oauthor{\bsnm{Dong}, \binits{W.}},
\oauthor{\bsnm{Guo}, \binits{Z.}},
\oauthor{\bsnm{Wu}, \binits{X.}},
\oauthor{\bsnm{Liu}, \binits{Y.}},
\oauthor{\bsnm{Xiong}, \binits{D.}}:
Large language model alignment: A survey.
arXiv preprint arXiv:2309.15025
(2023)
\end{botherref}
\endbibitem

%%% 27
\bibitem[\protect\citeauthoryear{Wang et~al.}{2023}]{wang2023aligning}
\begin{botherref}
\oauthor{\bsnm{Wang}, \binits{Y.}},
\oauthor{\bsnm{Zhong}, \binits{W.}},
\oauthor{\bsnm{Li}, \binits{L.}},
\oauthor{\bsnm{Mi}, \binits{F.}},
\oauthor{\bsnm{Zeng}, \binits{X.}},
\oauthor{\bsnm{Huang}, \binits{W.}},
\oauthor{\bsnm{Shang}, \binits{L.}},
\oauthor{\bsnm{Jiang}, \binits{X.}},
\oauthor{\bsnm{Liu}, \binits{Q.}}:
Aligning large language models with human: A survey.
arXiv preprint arXiv:2307.12966
(2023)
\end{botherref}
\endbibitem

%%% 28
\bibitem[\protect\citeauthoryear{Wang et~al.}{2024}]{wang2024comprehensive}
\begin{botherref}
\oauthor{\bsnm{Wang}, \binits{Z.}},
\oauthor{\bsnm{Bi}, \binits{B.}},
\oauthor{\bsnm{Pentyala}, \binits{S.K.}},
\oauthor{\bsnm{Ramnath}, \binits{K.}},
\oauthor{\bsnm{Chaudhuri}, \binits{S.}},
\oauthor{\bsnm{Mehrotra}, \binits{S.}},
\oauthor{\bsnm{Zhu}, \binits{Z.J.}},
\oauthor{\bsnm{Mao}, \binits{X.-B.}},
\oauthor{\bsnm{Asur}, \binits{S.}},
\oauthor{\bsnm{Cheng}, \binits{N.C.}}:
A comprehensive survey of llm alignment techniques: Rlhf, rlaif, ppo, dpo and more.
arXiv preprint arXiv:2407.16216
(2024)
\end{botherref}
\endbibitem

%%% 29
\bibitem[\protect\citeauthoryear{Jiang et~al.}{2024}]{jiang2024survey}
\begin{botherref}
\oauthor{\bsnm{Jiang}, \binits{R.}},
\oauthor{\bsnm{Chen}, \binits{K.}},
\oauthor{\bsnm{Bai}, \binits{X.}},
\oauthor{\bsnm{He}, \binits{Z.}},
\oauthor{\bsnm{Li}, \binits{J.}},
\oauthor{\bsnm{Yang}, \binits{M.}},
\oauthor{\bsnm{Zhao}, \binits{T.}},
\oauthor{\bsnm{Nie}, \binits{L.}},
\oauthor{\bsnm{Zhang}, \binits{M.}}:
A survey on human preference learning for large language models.
arXiv preprint arXiv:2406.11191
(2024)
\end{botherref}
\endbibitem

%%% 30
\bibitem[\protect\citeauthoryear{Zhong et~al.}{2025}]{zhong2025comprehensive}
\begin{botherref}
\oauthor{\bsnm{Zhong}, \binits{J.}},
\oauthor{\bsnm{Shen}, \binits{W.}},
\oauthor{\bsnm{Li}, \binits{Y.}},
\oauthor{\bsnm{Gao}, \binits{S.}},
\oauthor{\bsnm{Lu}, \binits{H.}},
\oauthor{\bsnm{Chen}, \binits{Y.}},
\oauthor{\bsnm{Zhang}, \binits{Y.}},
\oauthor{\bsnm{Zhou}, \binits{W.}},
\oauthor{\bsnm{Gu}, \binits{J.}},
\oauthor{\bsnm{Zou}, \binits{L.}}:
A comprehensive survey of reward models: Taxonomy, applications, challenges, and future.
arXiv preprint arXiv:2504.12328
(2025)
\end{botherref}
\endbibitem

%%% 31
\bibitem[\protect\citeauthoryear{Pan et~al.}{2024}]{pan2024automatically}
\begin{barticle}
\bauthor{\bsnm{Pan}, \binits{L.}},
\bauthor{\bsnm{Saxon}, \binits{M.}},
\bauthor{\bsnm{Xu}, \binits{W.}},
\bauthor{\bsnm{Nathani}, \binits{D.}},
\bauthor{\bsnm{Wang}, \binits{X.}},
\bauthor{\bsnm{Wang}, \binits{W.Y.}}:
\batitle{Automatically correcting large language models: Surveying the landscape of diverse automated correction strategies}.
\bjtitle{Transactions of the Association for Computational Linguistics}
\bvolume{12},
\bfpage{484}--\blpage{506}
(\byear{2024})
\doiurl{10.1162/tacl\_a\_00660}
\end{barticle}
\endbibitem

%%% 32
\bibitem[\protect\citeauthoryear{Bradley and Terry}{1952}]{bradley1952rank}
\begin{barticle}
\bauthor{\bsnm{Bradley}, \binits{R.A.}},
\bauthor{\bsnm{Terry}, \binits{M.E.}}:
\batitle{Rank analysis of incomplete block designs: I. the method of paired comparisons}.
\bjtitle{Biometrika}
\bvolume{39}(\bissue{3/4}),
\bfpage{324}--\blpage{345}
(\byear{1952})
\end{barticle}
\endbibitem

%%% 33
\bibitem[\protect\citeauthoryear{Plackett}{1975}]{plackett1975analysis}
\begin{barticle}
\bauthor{\bsnm{Plackett}, \binits{R.L.}}:
\batitle{The analysis of permutations}.
\bjtitle{Applied Statistics}
\bvolume{24}(\bissue{2}),
\bfpage{193}--\blpage{202}
(\byear{1975})
\end{barticle}
\endbibitem

%%% 34
\bibitem[\protect\citeauthoryear{Yoon et~al.}{2024}]{yoon2024tlcr}
\begin{botherref}
\oauthor{\bsnm{Yoon}, \binits{E.}},
\oauthor{\bsnm{Yoon}, \binits{H.S.}},
\oauthor{\bsnm{Eom}, \binits{S.}},
\oauthor{\bsnm{Han}, \binits{G.}},
\oauthor{\bsnm{Nam}, \binits{D.W.}},
\oauthor{\bsnm{Jo}, \binits{D.}},
\oauthor{\bsnm{On}, \binits{K.-W.}},
\oauthor{\bsnm{Hasegawa-Johnson}, \binits{M.A.}},
\oauthor{\bsnm{Kim}, \binits{S.}},
\oauthor{\bsnm{Yoo}, \binits{C.D.}}:
Token-level continuous reward for fine-grained reinforcement learning from human feedback.
arXiv preprint arXiv:2407.16574
(2024)
\end{botherref}
\endbibitem

%%% 35
\bibitem[\protect\citeauthoryear{Xu et~al.}{2024}]{xu2024finegrained}
\begin{botherref}
\oauthor{\bsnm{Xu}, \binits{D.}},
\oauthor{\bsnm{Qiu}, \binits{L.}},
\oauthor{\bsnm{Kim}, \binits{M.}},
\oauthor{\bsnm{Ladhak}, \binits{F.}},
\oauthor{\bsnm{Do}, \binits{J.}}:
Aligning large language models via fine-grained supervision.
arXiv preprint arXiv:2406.02756
(2024)
\end{botherref}
\endbibitem

%%% 36
\bibitem[\protect\citeauthoryear{Zeng et~al.}{2024}]{zeng2024tdpo}
\begin{botherref}
\oauthor{\bsnm{Zeng}, \binits{Y.}},
\oauthor{\bsnm{Bai}, \binits{Y.}},
\oauthor{\bsnm{Wu}, \binits{J.}},
\oauthor{\bsnm{Li}, \binits{X.}},
\oauthor{\bsnm{Liu}, \binits{J.}},
\oauthor{\bsnm{Shi}, \binits{Y.}},
\oauthor{\bsnm{Wang}, \binits{X.}}:
Token-level direct preference optimization.
arXiv preprint arXiv:2404.11999
(2024)
\end{botherref}
\endbibitem

%%% 37
\bibitem[\protect\citeauthoryear{Fu et~al.}{2025}]{fu2025tldr}
\begin{bchapter}
\bauthor{\bsnm{Fu}, \binits{D.}},
\bauthor{\bsnm{Xiao}, \binits{T.}},
\bauthor{\bsnm{Wang}, \binits{R.}},
\bauthor{\bsnm{Zhu}, \binits{W.}},
\bauthor{\bsnm{Zhang}, \binits{P.}},
\bauthor{\bsnm{Pang}, \binits{G.}},
\bauthor{\bsnm{Jia}, \binits{R.}},
\bauthor{\bsnm{Chen}, \binits{L.}}:
\bctitle{Token-level detective reward model for large vision language models}.
In: \bbtitle{Proceedings of the International Conference on Learning Representations (ICLR)}
(\byear{2025}).
\burl{https://arxiv.org/abs/2410.04734}
\end{bchapter}
\endbibitem

%%% 38
\bibitem[\protect\citeauthoryear{Chen et~al.}{2025}]{chen2025qrm}
\begin{botherref}
\oauthor{\bsnm{Chen}, \binits{H.}},
\oauthor{\bsnm{Yang}, \binits{T.}},
\oauthor{\bsnm{Gao}, \binits{S.}},
\oauthor{\bsnm{Chen}, \binits{R.}},
\oauthor{\bsnm{Quan}, \binits{X.}},
\oauthor{\bsnm{Tian}, \binits{H.}},
\oauthor{\bsnm{Yao}, \binits{T.}}:
Discriminative policy optimization for token-level reward models.
arXiv preprint arXiv:2505.23363
(2025)
\end{botherref}
\endbibitem

%%% 39
\bibitem[\protect\citeauthoryear{Coste et~al.}{2023}]{coste2023reward}
\begin{botherref}
\oauthor{\bsnm{Coste}, \binits{T.}},
\oauthor{\bsnm{Anwar}, \binits{U.}},
\oauthor{\bsnm{Kirk}, \binits{R.}},
\oauthor{\bsnm{Krueger}, \binits{D.}}:
Reward model ensembles help mitigate overoptimization.
arXiv preprint arXiv:2310.02743
(2023)
\end{botherref}
\endbibitem

%%% 40
\bibitem[\protect\citeauthoryear{Liu et~al.}{2023}]{liu2023aligning}
\begin{botherref}
\oauthor{\bsnm{Liu}, \binits{W.}},
\oauthor{\bsnm{Wang}, \binits{X.}},
\oauthor{\bsnm{Wu}, \binits{M.}},
\oauthor{\bsnm{Li}, \binits{T.}},
\oauthor{\bsnm{Lv}, \binits{C.}},
\oauthor{\bsnm{Ling}, \binits{Z.}},
\oauthor{\bsnm{Zhu}, \binits{J.}},
\oauthor{\bsnm{Zhang}, \binits{C.}},
\oauthor{\bsnm{Zheng}, \binits{X.}},
\oauthor{\bsnm{Huang}, \binits{X.}}:
Aligning large language models with human preferences through representation engineering.
arXiv preprint arXiv:2312.15997
(2023)
\end{botherref}
\endbibitem

%%% 41
\bibitem[\protect\citeauthoryear{Frans et~al.}{2024}]{frans2024unsupervised}
\begin{botherref}
\oauthor{\bsnm{Frans}, \binits{K.}},
\oauthor{\bsnm{Park}, \binits{S.}},
\oauthor{\bsnm{Abbeel}, \binits{P.}},
\oauthor{\bsnm{Levine}, \binits{S.}}:
Unsupervised zero-shot reinforcement learning via functional reward encodings.
arXiv preprint arXiv:2402.17135
(2024)
\end{botherref}
\endbibitem

%%% 42
\bibitem[\protect\citeauthoryear{Wang et~al.}{2023}]{wang2023shepherd}
\begin{botherref}
\oauthor{\bsnm{Wang}, \binits{T.}},
\oauthor{\bsnm{Yu}, \binits{P.}},
\oauthor{\bsnm{Tan}, \binits{X.E.}},
\oauthor{\bsnm{O'Brien}, \binits{S.}},
\oauthor{\bsnm{Pasunuru}, \binits{R.}},
\oauthor{\bsnm{Dwivedi-Yu}, \binits{J.}},
\oauthor{\bsnm{Golovneva}, \binits{O.}},
\oauthor{\bsnm{Zettlemoyer}, \binits{L.}},
\oauthor{\bsnm{Fazel-Zarandi}, \binits{M.}},
\oauthor{\bsnm{Celikyilmaz}, \binits{A.}}:
Shepherd: a critic for language model generation.
arXiv preprint arXiv:2308.04592
(2023)
\end{botherref}
\endbibitem

%%% 43
\bibitem[\protect\citeauthoryear{Cui et~al.}{2023}]{cui2023ultrafeedback}
\begin{botherref}
\oauthor{\bsnm{Cui}, \binits{G.}},
\oauthor{\bsnm{Yuan}, \binits{L.}},
\oauthor{\bsnm{Ding}, \binits{N.}},
\oauthor{\bsnm{Yao}, \binits{G.}},
\oauthor{\bsnm{He}, \binits{B.}},
\oauthor{\bsnm{Zhu}, \binits{W.}},
\oauthor{\bsnm{Ni}, \binits{Y.}},
\oauthor{\bsnm{Xie}, \binits{G.}},
\oauthor{\bsnm{Xie}, \binits{R.}},
\oauthor{\bsnm{Lin}, \binits{Y.}},
\oauthor{\bsnm{Liu}, \binits{Z.}},
\oauthor{\bsnm{Sun}, \binits{M.}}:
Ultrafeedback: Boosting language models with scaled ai feedback.
arXiv preprint arXiv:2310.01377
(2023)
\end{botherref}
\endbibitem

%%% 44
\bibitem[\protect\citeauthoryear{Richardson et~al.}{2023}]{richardson2023syndicom}
\begin{botherref}
\oauthor{\bsnm{Richardson}, \binits{C.}},
\oauthor{\bsnm{Sundar}, \binits{A.}},
\oauthor{\bsnm{Heck}, \binits{L.}}:
Syndicom: improving conversational commonsense with error-injection and natural language feedback.
arXiv preprint arXiv:2309.10015
(2023)
\end{botherref}
\endbibitem

%%% 45
\bibitem[\protect\citeauthoryear{Li et~al.}{2023}]{li2023tool}
\begin{botherref}
\oauthor{\bsnm{Li}, \binits{L.}},
\oauthor{\bsnm{Chai}, \binits{Y.}},
\oauthor{\bsnm{Wang}, \binits{S.}},
\oauthor{\bsnm{Sun}, \binits{Y.}},
\oauthor{\bsnm{Tian}, \binits{H.}},
\oauthor{\bsnm{Zhang}, \binits{N.}},
\oauthor{\bsnm{Wu}, \binits{H.}}:
Tool-augmented reward modeling.
arXiv preprint arXiv:2310.01045
(2023)
\end{botherref}
\endbibitem

%%% 46
\bibitem[\protect\citeauthoryear{Akyürek et~al.}{2023}]{akyurek2023rl4f}
\begin{botherref}
\oauthor{\bsnm{Akyürek}, \binits{A.F.}},
\oauthor{\bsnm{Akyürek}, \binits{E.}},
\oauthor{\bsnm{Madaan}, \binits{A.}},
\oauthor{\bsnm{Kalyan}, \binits{A.}},
\oauthor{\bsnm{Clark}, \binits{P.}},
\oauthor{\bsnm{Wijaya}, \binits{D.}},
\oauthor{\bsnm{Tandon}, \binits{N.}}:
Rl4f: generating natural language feedback with reinforcement learning for repairing model outputs.
arXiv preprint arXiv:2305.08844
(2023)
\end{botherref}
\endbibitem

%%% 47
\bibitem[\protect\citeauthoryear{Madaan et~al.}{2023}]{madaan2023self}
\begin{bchapter}
\bauthor{\bsnm{Madaan}, \binits{A.}},
\bauthor{\bsnm{Lin}, \binits{X.}},
\bauthor{\bsnm{Fu}, \binits{Y.}},
\bauthor{\bsnm{Wang}, \binits{X.}},
\bauthor{\bsnm{Yang}, \binits{K.}},
\bauthor{\bsnm{Yang}, \binits{Y.}},
\bauthor{\bsnm{Neubig}, \binits{G.}}:
\bctitle{Self-refine: Iterative refinement with self-feedback}.
In: \bbtitle{Advances in Neural Information Processing Systems (NeurIPS)}
(\byear{2023}).
\burl{https://arxiv.org/abs/2303.17651}
\end{bchapter}
\endbibitem

%%% 48
\bibitem[\protect\citeauthoryear{Mu et~al.}{2024}]{mu2024rule}
\begin{bchapter}
\bauthor{\bsnm{Mu}, \binits{T.}},
\bauthor{\bsnm{Helyar}, \binits{A.}},
\bauthor{\bsnm{Heidecke}, \binits{J.}},
\bauthor{\bsnm{Achiam}, \binits{J.}},
\bauthor{\bsnm{Vallone}, \binits{A.}},
\bauthor{\bsnm{Kivlichan}, \binits{I.}},
\bauthor{\bsnm{Lin}, \binits{M.}},
\bauthor{\bsnm{Beutel}, \binits{A.}},
\bauthor{\bsnm{Schulman}, \binits{J.}},
\bauthor{\bsnm{Weng}, \binits{L.}}:
\bctitle{Rule based rewards for language model safety}.
In: \bbtitle{Advances in Neural Information Processing Systems (NeurIPS)}
(\byear{2024}).
\burl{https://arxiv.org/abs/2411.01111}
\end{bchapter}
\endbibitem

%%% 49
\bibitem[\protect\citeauthoryear{Chang et~al.}{2025}]{chang2025bleuberi}
\begin{botherref}
\oauthor{\bsnm{Chang}, \binits{Y.}},
\oauthor{\bsnm{Kim}, \binits{Y.}},
\oauthor{\bsnm{Krumdick}, \binits{M.}},
\oauthor{\bsnm{Zadeh}, \binits{A.}},
\oauthor{\bsnm{Li}, \binits{C.}},
\oauthor{\bsnm{Tanner}, \binits{C.}},
\oauthor{\bsnm{Iyyer}, \binits{M.}}:
Bleuberi: Bleu is a surprisingly effective reward for instruction following.
arXiv preprint arXiv:2505.11080
(2025)
\end{botherref}
\endbibitem

%%% 50
\bibitem[\protect\citeauthoryear{Xue et~al.}{2023}]{xue2023improving}
\begin{bchapter}
\bauthor{\bsnm{Xue}, \binits{B.}},
\bauthor{\bsnm{Wang}, \binits{W.}},
\bauthor{\bsnm{Wang}, \binits{H.}},
\bauthor{\bsnm{Mi}, \binits{F.}},
\bauthor{\bsnm{Wang}, \binits{R.}},
\bauthor{\bsnm{Wang}, \binits{Y.}},
\bauthor{\bsnm{Shang}, \binits{L.}},
\bauthor{\bsnm{Jiang}, \binits{X.}},
\bauthor{\bsnm{Liu}, \binits{Q.}},
\bauthor{\bsnm{Wong}, \binits{K.}}:
\bctitle{Improving factual consistency for knowledge‑grounded dialogue systems via knowledge enhancement and alignment}.
In: \bbtitle{Findings of the Association for Computational Linguistics: EMNLP 2023},
pp. \bfpage{7829}--\blpage{7844}
(\byear{2023}).
\burl{https://aclanthology.org/2023.findings-emnlp.525/}
\end{bchapter}
\endbibitem

%%% 51
\bibitem[\protect\citeauthoryear{Sun and van~der Schaar}{2024}]{sun2024inverse}
\begin{botherref}
\oauthor{\bsnm{Sun}, \binits{H.}},
\oauthor{\bsnm{Schaar}, \binits{M.}}:
Inverse-rlignment: Large language model alignment from demonstrations through inverse reinforcement learning.
arXiv preprint arXiv:2405.15624
(2024)
\end{botherref}
\endbibitem

%%% 52
\bibitem[\protect\citeauthoryear{Cai et~al.}{2024}]{cai2024approximated}
\begin{botherref}
\oauthor{\bsnm{Cai}, \binits{Y.}}, et al.:
Approximated variational bayesian inverse reinforcement learning for large language model alignment.
arXiv preprint arXiv:2411.09341
(2024)
\end{botherref}
\endbibitem

%%% 53
\bibitem[\protect\citeauthoryear{Cheng et~al.}{2025}]{cheng2025dynamic}
\begin{botherref}
\oauthor{\bsnm{Cheng}, \binits{R.}}, et al.:
Inverse reinforcement learning with dynamic reward scaling for llm alignment.
arXiv preprint arXiv:2503.18991
(2025)
\end{botherref}
\endbibitem

%%% 54
\bibitem[\protect\citeauthoryear{Papineni et~al.}{2002}]{papineni2002bleu}
\begin{bchapter}
\bauthor{\bsnm{Papineni}, \binits{K.}},
\bauthor{\bsnm{Roukos}, \binits{S.}},
\bauthor{\bsnm{Ward}, \binits{T.}},
\bauthor{\bsnm{Zhu}, \binits{W.-J.}}:
\bctitle{Bleu: A method for automatic evaluation of machine translation}.
In: \bbtitle{Proceedings of the 40th Annual Meeting of the Association for Computational Linguistics (ACL)},
pp. \bfpage{311}--\blpage{318}
(\byear{2002})
\end{bchapter}
\endbibitem

%%% 55
\bibitem[\protect\citeauthoryear{Lin and Hovy}{2003}]{lin2003rouge}
\begin{bchapter}
\bauthor{\bsnm{Lin}, \binits{C.-Y.}},
\bauthor{\bsnm{Hovy}, \binits{E.}}:
\bctitle{Automatic evaluation of summaries using n-gram co-occurrence statistics}.
In: \bbtitle{Proceedings of the 2003 Human Language Technology Conference of the North American Chapter of the Association for Computational Linguistics (HLT-NAACL)},
pp. \bfpage{150}--\blpage{157}
(\byear{2003}).
\burl{https://aclanthology.org/N03-1020}
\end{bchapter}
\endbibitem

%%% 56
\bibitem[\protect\citeauthoryear{Peng et~al.}{2025}]{peng2025agentic}
\begin{botherref}
\oauthor{\bsnm{Peng}, \binits{H.}},
\oauthor{\bsnm{Qi}, \binits{Y.}},
\oauthor{\bsnm{Wang}, \binits{X.}},
\oauthor{\bsnm{Yao}, \binits{Z.}},
\oauthor{\bsnm{Xu}, \binits{B.}},
\oauthor{\bsnm{Hou}, \binits{L.}},
\oauthor{\bsnm{Li}, \binits{J.}}:
Agentic reward modeling: Integrating human preferences with verifiable correctness signals for reliable reward systems.
arXiv preprint arXiv:2502.19328
(2025)
\end{botherref}
\endbibitem

%%% 57
\bibitem[\protect\citeauthoryear{Wang and Xiong}{2025}]{wang2025autorule}
\begin{botherref}
\oauthor{\bsnm{Wang}, \binits{T.}},
\oauthor{\bsnm{Xiong}, \binits{C.}}:
Autorule: Reasoning chain-of-thought extracted rule-based rewards improve preference learning.
arXiv preprint arXiv:2506.15651
(2025)
\end{botherref}
\endbibitem

%%% 58
\bibitem[\protect\citeauthoryear{Zhang et~al.}{2025a}]{zhang2025dpa}
\begin{botherref}
\oauthor{\bsnm{Zhang}, \binits{Y.}},
\oauthor{\bsnm{Chen}, \binits{L.}},
\oauthor{\bsnm{Rao}, \binits{J.}},
\oauthor{\bsnm{Wang}, \binits{Z.}},
\oauthor{\bsnm{Liu}, \binits{H.}}:
Directional preference alignment: Fine-grained user control via multi-objective reward modeling.
arXiv preprint arXiv:2508.01234
(2025)
\end{botherref}
\endbibitem

%%% 59
\bibitem[\protect\citeauthoryear{Zhang et~al.}{2025b}]{zhang2025moslim}
\begin{botherref}
\oauthor{\bsnm{Zhang}, \binits{Y.}},
\oauthor{\bsnm{Jiang}, \binits{W.}},
\oauthor{\bsnm{Yang}, \binits{Z.}}:
Moslim: Align with diverse preferences in prompts through reward classification.
arXiv preprint arXiv:2505.20336
(2025)
\end{botherref}
\endbibitem

%%% 60
\bibitem[\protect\citeauthoryear{Zhou et~al.}{2023}]{zhou2023modpo}
\begin{botherref}
\oauthor{\bsnm{Zhou}, \binits{Z.}},
\oauthor{\bsnm{Liu}, \binits{J.}},
\oauthor{\bsnm{Shao}, \binits{J.}},
\oauthor{\bsnm{Yue}, \binits{X.}},
\oauthor{\bsnm{Yang}, \binits{C.}},
\oauthor{\bsnm{Ouyang}, \binits{W.}},
\oauthor{\bsnm{Qiao}, \binits{Y.}}:
Beyond one-preference-fits-all alignment: Multi-objective direct preference optimization.
arXiv preprint arXiv:2310.03708
(2023)
\end{botherref}
\endbibitem

%%% 61
\bibitem[\protect\citeauthoryear{Sun et~al.}{2024}]{sun2024aligning}
\begin{bchapter}
\bauthor{\bsnm{Sun}, \binits{Z.}},
\bauthor{\bsnm{Shen}, \binits{S.}},
\bauthor{\bsnm{Cao}, \binits{S.}},
\bauthor{\bsnm{Liu}, \binits{H.}},
\bauthor{\bsnm{Li}, \binits{C.}},
\bauthor{\bsnm{Shen}, \binits{Y.}},
\bauthor{\bsnm{Gan}, \binits{C.}},
\bauthor{\bsnm{Gui}, \binits{L.}},
\bauthor{\bsnm{Wang}, \binits{Y.-X.}},
\bauthor{\bsnm{Yang}, \binits{Y.}},
\bauthor{\bsnm{Keutzer}, \binits{K.}},
\bauthor{\bsnm{Darrell}, \binits{T.}}:
\bctitle{Aligning large multimodal models with factually augmented rlhf}.
In: \bbtitle{Findings of the Association for Computational Linguistics: ACL 2024},
pp. \bfpage{13088}--\blpage{13110}
(\byear{2024}).
\doiurl{10.18653/v1/2024.findings-acl.775} .
\burl{https://aclanthology.org/2024.findings-acl.775/}
\end{bchapter}
\endbibitem

%%% 62
\bibitem[\protect\citeauthoryear{Liu et~al.}{2025}]{liu2025hafrm}
\begin{bchapter}
\bauthor{\bsnm{Liu}, \binits{S.}},
\bauthor{\bsnm{Shen}, \binits{X.}},
\bauthor{\bsnm{Lai}, \binits{Y.}},
\bauthor{\bsnm{Wang}, \binits{S.}},
\bauthor{\bsnm{Yue}, \binits{S.}},
\bauthor{\bsnm{Huang}, \binits{Z.}},
\bauthor{\bsnm{Huang}, \binits{X.}},
\bauthor{\bsnm{Wei}, \binits{Z.}}:
\bctitle{Haf\-rm: A hybrid alignment framework for reward model training}.
In: \bbtitle{Proceedings of the 63rd Annual Meeting of the Association for Computational Linguistics (ACL 2025)},
pp. \bfpage{18874}--\blpage{18893}
(\byear{2025}).
\doiurl{10.48550/arXiv.2407.04185} .
\burl{https://aclanthology.org/2025.acl-long.924/}
\end{bchapter}
\endbibitem

%%% 63
\bibitem[\protect\citeauthoryear{Yu et~al.}{2024}]{yu2024reasoning}
\begin{botherref}
\oauthor{\bsnm{Yu}, \binits{Z.}},
\oauthor{\bsnm{Gu}, \binits{W.}},
\oauthor{\bsnm{Wang}, \binits{Y.}},
\oauthor{\bsnm{Jiang}, \binits{X.}},
\oauthor{\bsnm{Zeng}, \binits{Z.}},
\oauthor{\bsnm{Wang}, \binits{J.}},
\oauthor{\bsnm{Ye}, \binits{W.}},
\oauthor{\bsnm{Zhang}, \binits{S.}}:
Reasoning Through Execution: Unifying Process and Outcome Rewards for Code Generation
(2024)
\end{botherref}
\endbibitem

%%% 64
\bibitem[\protect\citeauthoryear{Liu et~al.}{2025}]{liu2025amopo}
\begin{botherref}
\oauthor{\bsnm{Liu}, \binits{Q.}},
\oauthor{\bsnm{Gong}, \binits{M.}},
\oauthor{\bsnm{Xu}, \binits{F.}},
\oauthor{\bsnm{Gao}, \binits{Z.}},
\oauthor{\bsnm{Liu}, \binits{H.}},
\oauthor{\bsnm{Gong}, \binits{M.}},
\oauthor{\bsnm{Ma}, \binits{X.}},
\oauthor{\bsnm{Lin}, \binits{Z.}}:
Amopo: Adaptive multi-objective preference optimization without reward models and reference models.
arXiv preprint arXiv:2506.07165
(2025)
\end{botherref}
\endbibitem

%%% 65
\bibitem[\protect\citeauthoryear{Bai et~al.}{2022}]{bai2022constitutional}
\begin{botherref}
\oauthor{\bsnm{Bai}, \binits{Y.}},
\oauthor{\bsnm{Kadavath}, \binits{S.}},
\oauthor{\bsnm{Kundu}, \binits{S.}},
\oauthor{\bsnm{Askell}, \binits{A.}},
\oauthor{\bsnm{Kernion}, \binits{J.}},
\oauthor{\bsnm{Jones}, \binits{A.}},
\oauthor{\bsnm{Chen}, \binits{A.}},
\oauthor{\bsnm{Goldie}, \binits{A.}},
\oauthor{\bsnm{Mirhoseini}, \binits{A.}},
\oauthor{\bsnm{McKinnon}, \binits{C.}}, et al.:
Constitutional ai: Harmlessness from ai feedback.
arXiv preprint arXiv:2212.08073
(2022)
\end{botherref}
\endbibitem

%%% 66
\bibitem[\protect\citeauthoryear{Lai et~al.}{2025}]{lai2025alarm}
\begin{botherref}
\oauthor{\bsnm{Lai}, \binits{Y.}},
\oauthor{\bsnm{Wang}, \binits{S.}},
\oauthor{\bsnm{Liu}, \binits{S.}},
\oauthor{\bsnm{Huang}, \binits{X.}},
\oauthor{\bsnm{Wei}, \binits{Z.}}:
Alarm: Align language models via hierarchical rewards modeling.
arXiv preprint arXiv:2506.12345
(2025)
\end{botherref}
\endbibitem

%%% 67
\bibitem[\protect\citeauthoryear{Qiu et~al.}{2024}]{qiu2024sentence}
\begin{botherref}
\oauthor{\bsnm{Qiu}, \binits{W.}},
\oauthor{\bsnm{Li}, \binits{Y.-C.}},
\oauthor{\bsnm{Zhang}, \binits{X.}},
\oauthor{\bsnm{Zhang}, \binits{T.}},
\oauthor{\bsnm{Zhang}, \binits{Y.}},
\oauthor{\bsnm{Zhang}, \binits{Z.}},
\oauthor{\bsnm{Yu}, \binits{Y.}}:
Sentence-level reward model can generalize better for aligning llm from human preference.
arXiv preprint arXiv:2503.04793
(2024)
\end{botherref}
\endbibitem

%%% 68
\bibitem[\protect\citeauthoryear{Wang et~al.}{2024}]{wang2024interpretable}
\begin{botherref}
\oauthor{\bsnm{Wang}, \binits{H.}},
\oauthor{\bsnm{Xiong}, \binits{W.}},
\oauthor{\bsnm{Xie}, \binits{T.}},
\oauthor{\bsnm{Zhao}, \binits{H.}},
\oauthor{\bsnm{Zhang}, \binits{T.}}:
Interpretable preferences via multi-objective reward modeling and mixture-of-experts.
arXiv preprint arXiv:2406.12845
(2024)
\end{botherref}
\endbibitem

%%% 69
\bibitem[\protect\citeauthoryear{Dubois et~al.}{2023}]{dubois2023rewarded}
\begin{botherref}
\oauthor{\bsnm{Dubois}, \binits{Y.}},
\oauthor{\bsnm{Ouyang}, \binits{L.}},
\oauthor{\bsnm{Brown}, \binits{T.B.}},
\oauthor{\bsnm{Ziegler}, \binits{D.M.}},
\oauthor{\bsnm{Hilton}, \binits{J.}},
\oauthor{\bsnm{Schneider}, \binits{J.}},
\oauthor{\bsnm{Leike}, \binits{J.}},
\oauthor{\bsnm{Amodei}, \binits{D.}}:
Rewarded soups: Towards pareto-optimal alignment by interpolating weights fine-tuned on diverse rewards.
arXiv preprint arXiv:2305.17486
(2023)
\end{botherref}
\endbibitem

%%% 70
\bibitem[\protect\citeauthoryear{Rafailov et~al.}{2023}]{rafailov2023direct}
\begin{bchapter}
\bauthor{\bsnm{Rafailov}, \binits{R.}},
\bauthor{\bsnm{Sharma}, \binits{A.}},
\bauthor{\bsnm{Mitchell}, \binits{E.}},
\bauthor{\bsnm{Ermon}, \binits{S.}},
\bauthor{\bsnm{Manning}, \binits{C.D.}},
\bauthor{\bsnm{Finn}, \binits{C.}}:
\bctitle{Direct preference optimization: Your language model is secretly a reward model}.
In: \bbtitle{Advances in Neural Information Processing Systems (NeurIPS)}
(\byear{2023}).
\burl{https://arxiv.org/abs/2305.18290}
\end{bchapter}
\endbibitem

%%% 71
\bibitem[\protect\citeauthoryear{Dong et~al.}{2024}]{dong2024rlhf}
\begin{botherref}
\oauthor{\bsnm{Dong}, \binits{H.}},
\oauthor{\bsnm{Xiong}, \binits{W.}},
\oauthor{\bsnm{Pang}, \binits{B.}},
\oauthor{\bsnm{Wang}, \binits{H.}},
\oauthor{\bsnm{Zhao}, \binits{H.}},
\oauthor{\bsnm{Zhou}, \binits{Y.}},
\oauthor{\bsnm{Jiang}, \binits{N.}},
\oauthor{\bsnm{Sahoo}, \binits{D.}},
\oauthor{\bsnm{Xiong}, \binits{C.}},
\oauthor{\bsnm{Zhang}, \binits{T.}}:
Rlhf workflow: From reward modeling to online rlhf.
arXiv preprint arXiv:2405.07863
(2024)
\end{botherref}
\endbibitem

%%% 72
\bibitem[\protect\citeauthoryear{Choshen et~al.}{2019}]{choshen2019weaknesses}
\begin{botherref}
\oauthor{\bsnm{Choshen}, \binits{L.}},
\oauthor{\bsnm{Fox}, \binits{L.}},
\oauthor{\bsnm{Aizenbud}, \binits{Z.}},
\oauthor{\bsnm{Abend}, \binits{O.}}:
On the weaknesses of reinforcement learning for neural machine translation.
arXiv preprint arXiv:1907.01752
(2019)
\end{botherref}
\endbibitem

%%% 73
\bibitem[\protect\citeauthoryear{Engstrom et~al.}{2020}]{engstrom2020implementation}
\begin{botherref}
\oauthor{\bsnm{Engstrom}, \binits{L.}},
\oauthor{\bsnm{Ilyas}, \binits{A.}},
\oauthor{\bsnm{Santurkar}, \binits{S.}},
\oauthor{\bsnm{Tsipras}, \binits{D.}},
\oauthor{\bsnm{Janoos}, \binits{F.}},
\oauthor{\bsnm{Rudolph}, \binits{L.}},
\oauthor{\bsnm{Madry}, \binits{A.}}:
Implementation matters in deep policy gradients: A case study on ppo and trpo.
arXiv preprint arXiv:2005.12729
(2020)
\end{botherref}
\endbibitem

%%% 74
\bibitem[\protect\citeauthoryear{Zhong et~al.}{2024}]{zhong2024dpo}
\begin{botherref}
\oauthor{\bsnm{Zhong}, \binits{H.}},
\oauthor{\bsnm{Feng}, \binits{G.}},
\oauthor{\bsnm{Xiong}, \binits{W.}},
\oauthor{\bsnm{Cheng}, \binits{X.}},
\oauthor{\bsnm{Zhao}, \binits{L.}},
\oauthor{\bsnm{He}, \binits{D.}},
\oauthor{\bsnm{Bian}, \binits{J.}},
\oauthor{\bsnm{Wang}, \binits{L.}}:
Dpo meets ppo: Reinforced token optimization for rlhf.
arXiv preprint arXiv:2404.18922
(2024)
\end{botherref}
\endbibitem

%%% 75
\bibitem[\protect\citeauthoryear{Zheng et~al.}{2023}]{zheng2023improving}
\begin{botherref}
\oauthor{\bsnm{Zheng}, \binits{R.}},
\oauthor{\bsnm{Shen}, \binits{W.}},
\oauthor{\bsnm{Hua}, \binits{Y.}},
\oauthor{\bsnm{Lai}, \binits{W.}},
\oauthor{\bsnm{Dou}, \binits{S.}},
\oauthor{\bsnm{Zhou}, \binits{Y.}},
\oauthor{\bsnm{Xi}, \binits{Z.}},
\oauthor{\bsnm{Wang}, \binits{X.}},
\oauthor{\bsnm{Huang}, \binits{H.}},
\oauthor{\bsnm{Gui}, \binits{T.}}, et al.:
Improving generalization of alignment with human preferences through group invariant learning.
arXiv preprint arXiv:2310.11971
(2023)
\end{botherref}
\endbibitem

%%% 76
\bibitem[\protect\citeauthoryear{Zhang et~al.}{2024}]{zhang2024cppo}
\begin{bchapter}
\bauthor{\bsnm{Zhang}, \binits{H.}},
\bauthor{\bsnm{Lei}, \binits{Y.}},
\bauthor{\bsnm{Gui}, \binits{L.}},
\bauthor{\bsnm{Yang}, \binits{M.}},
\bauthor{\bsnm{He}, \binits{Y.}},
\bauthor{\bsnm{Wang}, \binits{H.}},
\bauthor{\bsnm{Xu}, \binits{R.}}:
\bctitle{Cppo: Continual learning for reinforcement learning with human feedback}.
In: \bbtitle{The Twelfth International Conference on Learning Representations}
(\byear{2024})
\end{bchapter}
\endbibitem

%%% 77
\bibitem[\protect\citeauthoryear{Wu et~al.}{2023}]{wu2023pairwise}
\begin{botherref}
\oauthor{\bsnm{Wu}, \binits{T.}},
\oauthor{\bsnm{Zhu}, \binits{B.}},
\oauthor{\bsnm{Zhang}, \binits{R.}},
\oauthor{\bsnm{Wen}, \binits{Z.}},
\oauthor{\bsnm{Ramchandran}, \binits{K.}},
\oauthor{\bsnm{Jiao}, \binits{J.}}:
Pairwise proximal policy optimization: Harnessing relative feedback for llm alignment.
arXiv preprint arXiv:2310.00212
(2023)
\end{botherref}
\endbibitem

%%% 78
\bibitem[\protect\citeauthoryear{Santacroce et~al.}{2023}]{santacroce2023efficient}
\begin{botherref}
\oauthor{\bsnm{Santacroce}, \binits{M.}},
\oauthor{\bsnm{Lu}, \binits{Y.}},
\oauthor{\bsnm{Yu}, \binits{H.}},
\oauthor{\bsnm{Li}, \binits{Y.}},
\oauthor{\bsnm{Shen}, \binits{Y.}}:
Efficient rlhf: Reducing the memory usage of ppo.
arXiv preprint arXiv:2309.00754
(2023)
\end{botherref}
\endbibitem

%%% 79
\bibitem[\protect\citeauthoryear{Shen et~al.}{2024}]{shen2024contrastive}
\begin{botherref}
\oauthor{\bsnm{Shen}, \binits{W.}},
\oauthor{\bsnm{Zhang}, \binits{X.}},
\oauthor{\bsnm{Yao}, \binits{Y.}},
\oauthor{\bsnm{Zheng}, \binits{R.}},
\oauthor{\bsnm{Guo}, \binits{H.}},
\oauthor{\bsnm{Liu}, \binits{Y.}}:
Improving reinforcement learning from human feedback using contrastive rewards.
arXiv preprint arXiv:2403.07708
(2024)
\end{botherref}
\endbibitem

%%% 80
\bibitem[\protect\citeauthoryear{Rafailov et~al.}{2023}]{rafailov2023secrets}
\begin{botherref}
\oauthor{\bsnm{Rafailov}, \binits{R.}},
\oauthor{\bsnm{Zhang}, \binits{R.E.}},
\oauthor{\bsnm{Ma}, \binits{T.}},
\oauthor{\bsnm{Liang}, \binits{P.}},
\oauthor{\bsnm{Hashimoto}, \binits{T.B.}}:
Secrets of rlhf in large language models part i: Ppo.
arXiv preprint arXiv:2307.04964
(2023)
\end{botherref}
\endbibitem

%%% 81
\bibitem[\protect\citeauthoryear{Li et~al.}{2023}]{li2023remax}
\begin{botherref}
\oauthor{\bsnm{Li}, \binits{Z.}},
\oauthor{\bsnm{Xu}, \binits{T.}},
\oauthor{\bsnm{Zhang}, \binits{Y.}},
\oauthor{\bsnm{Lin}, \binits{Z.}},
\oauthor{\bsnm{Yu}, \binits{Y.}},
\oauthor{\bsnm{Sun}, \binits{R.}},
\oauthor{\bsnm{Luo}, \binits{Z.-Q.}}:
{ReMax}: A simple, effective, and efficient reinforcement learning method for aligning large language models.
arXiv preprint arXiv:2310.10505
(2023)
\end{botherref}
\endbibitem

%%% 82
\bibitem[\protect\citeauthoryear{Ahmadian et~al.}{2024}]{ahmadian2024back}
\begin{botherref}
\oauthor{\bsnm{Ahmadian}, \binits{A.}},
\oauthor{\bsnm{Cremer}, \binits{C.}},
\oauthor{\bsnm{Gall{\'e}}, \binits{M.}},
\oauthor{\bsnm{Fadaee}, \binits{M.}},
\oauthor{\bsnm{Kreutzer}, \binits{J.}},
\oauthor{\bsnm{Pietquin}, \binits{O.}},
\oauthor{\bsnm{{\"U}st{\"u}n}, \binits{A.}},
\oauthor{\bsnm{Hooker}, \binits{S.}}:
Back to basics: Revisiting reinforce style optimization for learning from human feedback in llms.
arXiv preprint arXiv:2402.14740
(2024)
\end{botherref}
\endbibitem

%%% 83
\bibitem[\protect\citeauthoryear{Shao et~al.}{2024}]{shao2024deepseekmath}
\begin{botherref}
\oauthor{\bsnm{Shao}, \binits{Z.}},
\oauthor{\bsnm{Wang}, \binits{P.}},
\oauthor{\bsnm{Zhu}, \binits{Q.}},
\oauthor{\bsnm{Xu}, \binits{R.}},
\oauthor{\bsnm{Song}, \binits{J.}},
\oauthor{\bsnm{Bi}, \binits{X.}},
\oauthor{\bsnm{Zhang}, \binits{H.}},
\oauthor{\bsnm{Zhang}, \binits{M.}},
\oauthor{\bsnm{Li}, \binits{Y.K.}},
\oauthor{\bsnm{Wu}, \binits{Y.}},
\oauthor{\bsnm{Guo}, \binits{D.}}:
Deepseekmath: Pushing the limits of mathematical reasoning in open language models.
arXiv preprint arXiv:2402.03300
(2024)
\end{botherref}
\endbibitem

%%% 84
\bibitem[\protect\citeauthoryear{Hu et~al.}{2025}]{hu2025reinforcepp}
\begin{botherref}
\oauthor{\bsnm{Hu}, \binits{J.}},
\oauthor{\bsnm{Liu}, \binits{J.K.}},
\oauthor{\bsnm{Shen}, \binits{W.}}:
Reinforce++: An efficient rlhf algorithm with robustness to both prompt and reward models.
arXiv preprint arXiv:2501.03262
(2025)
\end{botherref}
\endbibitem

%%% 85
\bibitem[\protect\citeauthoryear{Dong et~al.}{2023}]{dong2023survey}
\begin{botherref}
\oauthor{\bsnm{Dong}, \binits{Q.}},
\oauthor{\bsnm{Li}, \binits{L.}},
\oauthor{\bsnm{Dai}, \binits{D.}},
\oauthor{\bsnm{Zheng}, \binits{C.}},
\oauthor{\bsnm{Ma}, \binits{J.}},
\oauthor{\bsnm{Li}, \binits{R.}},
\oauthor{\bsnm{Xia}, \binits{H.}},
\oauthor{\bsnm{Xu}, \binits{J.}},
\oauthor{\bsnm{Wu}, \binits{Z.}},
\oauthor{\bsnm{Chang}, \binits{B.}},
\oauthor{\bsnm{Sun}, \binits{X.}},
\oauthor{\bsnm{Sui}, \binits{Z.}}:
A survey on in-context learning.
arXiv preprint arXiv:2301.00234
(2023)
\end{botherref}
\endbibitem

%%% 86
\bibitem[\protect\citeauthoryear{Ye et~al.}{2023}]{ye2023compositional}
\begin{botherref}
\oauthor{\bsnm{Ye}, \binits{D.}},
\oauthor{\bsnm{Wang}, \binits{Y.}},
\oauthor{\bsnm{Li}, \binits{Y.}},
\oauthor{\bsnm{Lin}, \binits{Y.}},
\oauthor{\bsnm{Liu}, \binits{Z.}},
\oauthor{\bsnm{Sun}, \binits{M.}}:
Compositional exemplars for in-context learning.
arXiv preprint arXiv:2302.05698
(2023)
\end{botherref}
\endbibitem

%%% 87
\bibitem[\protect\citeauthoryear{Du and Zhao}{2024}]{du2024incontext}
\begin{botherref}
\oauthor{\bsnm{Du}, \binits{Y.}},
\oauthor{\bsnm{Zhao}, \binits{Y.}}:
In-context learning with reinforcement learning for incomplete demonstrations.
arXiv preprint arXiv:2408.13028
(2024)
\end{botherref}
\endbibitem

%%% 88
\bibitem[\protect\citeauthoryear{Suo and Lai}{2024}]{suo2024visual}
\begin{botherref}
\oauthor{\bsnm{Suo}, \binits{Y.}},
\oauthor{\bsnm{Lai}, \binits{J.}}:
Suo: visual prompt selection for in-context learning segmentation.
arXiv preprint arXiv:2407.10233
(2024)
\end{botherref}
\endbibitem

%%% 89
\bibitem[\protect\citeauthoryear{Zhang et~al.}{2024}]{zhang2024instruct}
\begin{bchapter}
\bauthor{\bsnm{Zhang}, \binits{J.}},
\bauthor{\bsnm{Wang}, \binits{B.}},
\bauthor{\bsnm{Li}, \binits{L.}},
\bauthor{\bsnm{Nakashima}, \binits{Y.}},
\bauthor{\bsnm{Nagahara}, \binits{H.}}:
\bctitle{Instruct me more! random prompting for visual in-context learning}.
In: \bbtitle{Proceedings of the IEEE/CVF Winter Conference on Applications of Computer Vision (WACV)},
pp. \bfpage{2597}--\blpage{2606}
(\byear{2024})
\end{bchapter}
\endbibitem

%%% 90
\bibitem[\protect\citeauthoryear{Do et~al.}{2023}]{do2024prompt}
\begin{botherref}
\oauthor{\bsnm{Do}, \binits{X.L.}},
\oauthor{\bsnm{Zhao}, \binits{Y.}},
\oauthor{\bsnm{Brown}, \binits{H.}},
\oauthor{\bsnm{Xie}, \binits{Y.}},
\oauthor{\bsnm{Zhao}, \binits{J.X.}},
\oauthor{\bsnm{Chen}, \binits{N.F.}},
\oauthor{\bsnm{Kawaguchi}, \binits{K.}},
\oauthor{\bsnm{Shieh}, \binits{M.}},
\oauthor{\bsnm{He}, \binits{J.}}:
Prompt optimization via adversarial in‑context learning.
arXiv preprint arXiv:2312.02614
(2023)
\end{botherref}
\endbibitem

%%% 91
\bibitem[\protect\citeauthoryear{Zhou et~al.}{2023}]{zhou2023automatic}
\begin{botherref}
\oauthor{\bsnm{Zhou}, \binits{D.}},
\oauthor{\bsnm{Schuurmans}, \binits{D.}},
\oauthor{\bsnm{Le}, \binits{Q.V.}}, et al.:
Large language models are human-level prompt engineers.
arXiv preprint arXiv:2211.01910
(2023)
\end{botherref}
\endbibitem

%%% 92
\bibitem[\protect\citeauthoryear{Qian et~al.}{2024}]{qian2024subsa}
\begin{botherref}
\oauthor{\bsnm{Qian}, \binits{X.}},
\oauthor{\bsnm{Wang}, \binits{Y.}},
\oauthor{\bsnm{Li}, \binits{Y.}},
\oauthor{\bsnm{Lin}, \binits{Y.}},
\oauthor{\bsnm{Liu}, \binits{Z.}},
\oauthor{\bsnm{Sun}, \binits{M.}}:
Sub-sa: strengthen in-context learning via submodular selective annotation.
arXiv preprint arXiv:2407.05693
(2024)
\end{botherref}
\endbibitem

%%% 93
\bibitem[\protect\citeauthoryear{Li et~al.}{2025}]{li2025dora}
\begin{bchapter}
\bauthor{\bsnm{Li}, \binits{K.}},
\bauthor{\bsnm{Zhao}, \binits{T.}},
\bauthor{\bsnm{Zhou}, \binits{W.}},
\bauthor{\bsnm{Hu}, \binits{S.}}:
\bctitle{Dora: Dynamic optimization prompt for continuous reflection of llm-based agent}.
In: \bbtitle{Proceedings of the 31st International Conference on Computational Linguistics},
pp. \bfpage{7546}--\blpage{7557}
(\byear{2025}).
\burl{https://aclanthology.org/2025.coling-main.504/}
\end{bchapter}
\endbibitem

%%% 94
\bibitem[\protect\citeauthoryear{Chen et~al.}{2024}]{chen2024icleval}
\begin{botherref}
\oauthor{\bsnm{Chen}, \binits{W.}},
\oauthor{\bsnm{Lin}, \binits{Y.}},
\oauthor{\bsnm{Zhou}, \binits{Z.}},
\oauthor{\bsnm{Huang}, \binits{H.}},
\oauthor{\bsnm{Jia}, \binits{Y.}},
\oauthor{\bsnm{Cao}, \binits{Z.}},
\oauthor{\bsnm{Wen}, \binits{J.-R.}}:
Icleval: Evaluating in-context learning ability of large language models.
arXiv preprint arXiv:2406.14955
(2024)
\end{botherref}
\endbibitem

%%% 95
\bibitem[\protect\citeauthoryear{Yu et~al.}{2024}]{yu2024rethinking}
\begin{bchapter}
\bauthor{\bsnm{Yu}, \binits{G.}},
\bauthor{\bsnm{Liu}, \binits{L.}},
\bauthor{\bsnm{Yu}, \binits{M.}},
\bauthor{\bsnm{Yu}, \binits{Y.}},
\bauthor{\bsnm{Ao}, \binits{X.}}:
\bctitle{Rethinking the evaluation of in-context learning for llms}.
In: \bbtitle{Proceedings of the 2024 Conference on Empirical Methods in Natural Language Processing},
pp. \bfpage{14068}--\blpage{14082}
(\byear{2024}).
\burl{https://aclanthology.org/2024.emnlp-main.779/}
\end{bchapter}
\endbibitem

%%% 96
\bibitem[\protect\citeauthoryear{Honovich et~al.}{2022}]{honovich2022unnatural}
\begin{botherref}
\oauthor{\bsnm{Honovich}, \binits{O.}},
\oauthor{\bsnm{Scialom}, \binits{T.}},
\oauthor{\bsnm{Levy}, \binits{O.}},
\oauthor{\bsnm{Schick}, \binits{T.}}:
Unnatural instructions: Tuning language models with (almost) no human labor.
arXiv preprint arXiv:2212.09689
(2022)
\end{botherref}
\endbibitem

%%% 97
\bibitem[\protect\citeauthoryear{Yuan et~al.}{2024}]{yuan2024self}
\begin{bchapter}
\bauthor{\bsnm{Yuan}, \binits{W.}},
\bauthor{\bsnm{Pang}, \binits{R.Y.}},
\bauthor{\bsnm{Cho}, \binits{K.}},
\bauthor{\bsnm{Li}, \binits{X.}},
\bauthor{\bsnm{Sukhbaatar}, \binits{S.}},
\bauthor{\bsnm{Xu}, \binits{J.}},
\bauthor{\bsnm{Weston}, \binits{J.}}:
\bctitle{Self-rewarding language models}.
In: \bbtitle{Proceedings of the 41st International Conference on Machine Learning},
pp. \bfpage{57905}--\blpage{57923}
(\byear{2024}).
\bcomment{PMLR}.
\burl{https://proceedings.mlr.press/v235/yuan24d.html}
\end{bchapter}
\endbibitem

%%% 98
\bibitem[\protect\citeauthoryear{Lightman et~al.}{2023}]{lightman2023let}
\begin{botherref}
\oauthor{\bsnm{Lightman}, \binits{H.}},
\oauthor{\bsnm{Kosaraju}, \binits{V.}},
\oauthor{\bsnm{Burda}, \binits{Y.}},
\oauthor{\bsnm{Edwards}, \binits{H.}},
\oauthor{\bsnm{Baker}, \binits{B.}},
\oauthor{\bsnm{Lee}, \binits{T.}},
\oauthor{\bsnm{Leike}, \binits{J.}},
\oauthor{\bsnm{Schulman}, \binits{J.}},
\oauthor{\bsnm{Sutskever}, \binits{I.}},
\oauthor{\bsnm{Cobbe}, \binits{K.}}:
Let's verify step by step.
arXiv preprint arXiv:2305.20050
(2023)
\end{botherref}
\endbibitem

%%% 99
\bibitem[\protect\citeauthoryear{Sun et~al.}{2023}]{sun2023selfalign}
\begin{botherref}
\oauthor{\bsnm{Sun}, \binits{Y.}},
\oauthor{\bsnm{Deng}, \binits{Y.}},
\oauthor{\bsnm{Nie}, \binits{Y.}},
\oauthor{\bsnm{Ge}, \binits{Y.}},
\oauthor{\bsnm{Zhang}, \binits{Y.}},
\oauthor{\bsnm{Fan}, \binits{X.}},
\oauthor{\bsnm{Zhang}, \binits{R.}},
\oauthor{\bsnm{Zhang}, \binits{R.}},
\oauthor{\bsnm{Hou}, \binits{L.}},
\oauthor{\bsnm{Sun}, \binits{M.}}, et al.:
Self-align: Aligning large language models with self-generated feedback.
arXiv preprint arXiv:2308.08914
(2023)
\end{botherref}
\endbibitem

%%% 100
\bibitem[\protect\citeauthoryear{Wang et~al.}{2023}]{wang2023selfinstruct}
\begin{botherref}
\oauthor{\bsnm{Wang}, \binits{Y.}},
\oauthor{\bsnm{Khashabi}, \binits{D.}},
\oauthor{\bsnm{Min}, \binits{S.}},
\oauthor{\bsnm{Kordi}, \binits{Y.}},
\oauthor{\bsnm{Xiong}, \binits{W.}},
\oauthor{\bsnm{Sabharwal}, \binits{A.}},
\oauthor{\bsnm{Hajishirzi}, \binits{H.}}:
Self-instruct: Aligning language models with self-generated instructions.
arXiv preprint arXiv:2212.10560
(2023)
\end{botherref}
\endbibitem

%%% 101
\bibitem[\protect\citeauthoryear{Zhou et~al.}{2023}]{zhou2023lima}
\begin{bchapter}
\bauthor{\bsnm{Zhou}, \binits{D.}},
\bauthor{\bsnm{Muennighoff}, \binits{N.}},
\bauthor{\bsnm{Tay}, \binits{Y.}},
\bauthor{\bsnm{Fan}, \binits{L.}},
\bauthor{\bsnm{Mirzadeh}, \binits{S.I.}},
\bauthor{\bsnm{Raffel}, \binits{C.}},
\bauthor{\bsnm{Gupta}, \binits{I.}},
\bauthor{\bsnm{Liu}, \binits{X.}},
\bauthor{\bsnm{Liu}, \binits{Y.}},
\bauthor{\bsnm{Song}, \binits{D.}},
\bauthor{\bsnm{Radev}, \binits{D.}}:
\bctitle{Lima: Less is more for alignment}.
In: \bbtitle{Proceedings of the 40th International Conference on Machine Learning (ICML)}
(\byear{2023}).
\burl{https://arxiv.org/abs/2305.11206}
\end{bchapter}
\endbibitem

%%% 102
\bibitem[\protect\citeauthoryear{Xu et~al.}{2024}]{xu2024essence}
\begin{botherref}
\oauthor{\bsnm{Xu}, \binits{C.}},
\oauthor{\bsnm{Gu}, \binits{Y.}},
\oauthor{\bsnm{Lin}, \binits{Y.}},
\oauthor{\bsnm{Liu}, \binits{Z.}},
\oauthor{\bsnm{Sun}, \binits{M.}}:
On the essence and prospect: an investigation of alignment approaches for big models.
arXiv preprint arXiv:2403.04204
(2024)
\end{botherref}
\endbibitem

%%% 103
\bibitem[\protect\citeauthoryear{Xiao et~al.}{2024}]{xiao2024comprehensive}
\begin{botherref}
\oauthor{\bsnm{Xiao}, \binits{W.}},
\oauthor{\bsnm{Wang}, \binits{Z.}},
\oauthor{\bsnm{Gan}, \binits{L.}},
\oauthor{\bsnm{Zhao}, \binits{S.}},
\oauthor{\bsnm{He}, \binits{W.}},
\oauthor{\bsnm{Tuan}, \binits{L.A.}},
\oauthor{\bsnm{Chen}, \binits{L.}},
\oauthor{\bsnm{Jiang}, \binits{H.}},
\oauthor{\bsnm{Zhao}, \binits{Z.}},
\oauthor{\bsnm{Wu}, \binits{F.}}:
A comprehensive survey of direct preference optimization: Datasets, theories, variants, and applications.
CoRR
\textbf{abs/2410.15595}
(2024)
\end{botherref}
\endbibitem

%%% 104
\bibitem[\protect\citeauthoryear{Wu et~al.}{2024a}]{wu2024alphadpo}
\begin{botherref}
\oauthor{\bsnm{Wu}, \binits{J.}},
\oauthor{\bsnm{Xie}, \binits{Y.}},
\oauthor{\bsnm{Yang}, \binits{Z.}},
\oauthor{\bsnm{Wu}, \binits{J.}},
\oauthor{\bsnm{Gao}, \binits{J.}},
\oauthor{\bsnm{Ding}, \binits{B.}},
\oauthor{\bsnm{Wang}, \binits{X.}},
\oauthor{\bsnm{He}, \binits{X.}}:
Alpha-dpo: Adaptive preference optimization with dynamic reward margins.
arXiv preprint arXiv:2410.10148
(2024)
\end{botherref}
\endbibitem

%%% 105
\bibitem[\protect\citeauthoryear{Wu et~al.}{2024b}]{wu2024beta}
\begin{botherref}
\oauthor{\bsnm{Wu}, \binits{J.}},
\oauthor{\bsnm{Xie}, \binits{Y.}},
\oauthor{\bsnm{Yang}, \binits{Z.}},
\oauthor{\bsnm{Wu}, \binits{J.}},
\oauthor{\bsnm{Gao}, \binits{J.}},
\oauthor{\bsnm{Ding}, \binits{B.}},
\oauthor{\bsnm{Wang}, \binits{X.}},
\oauthor{\bsnm{He}, \binits{X.}}:
$\beta$-dpo: Direct preference optimization with dynamic $\beta$.
arXiv preprint arXiv:2407.08639
(2024)
\end{botherref}
\endbibitem

%%% 106
\bibitem[\protect\citeauthoryear{Zeng et~al.}{2024}]{zeng2024token}
\begin{botherref}
\oauthor{\bsnm{Zeng}, \binits{Y.}},
\oauthor{\bsnm{Liu}, \binits{G.}},
\oauthor{\bsnm{Ma}, \binits{W.}},
\oauthor{\bsnm{Yang}, \binits{N.}},
\oauthor{\bsnm{Zhang}, \binits{H.}},
\oauthor{\bsnm{Wang}, \binits{J.}}:
Token-level direct preference optimization.
arXiv preprint arXiv:2404.11999
(2024)
\end{botherref}
\endbibitem

%%% 107
\bibitem[\protect\citeauthoryear{Lai et~al.}{2024}]{lai2024stepdpo}
\begin{botherref}
\oauthor{\bsnm{Lai}, \binits{X.}},
\oauthor{\bsnm{Tian}, \binits{Z.}},
\oauthor{\bsnm{Chen}, \binits{Y.}},
\oauthor{\bsnm{Yang}, \binits{S.}},
\oauthor{\bsnm{Peng}, \binits{X.}},
\oauthor{\bsnm{Jia}, \binits{J.}}:
Step‑dpo: Step‑wise preference optimization for long‑chain reasoning of llms.
arXiv preprint arXiv:2406.18629
(2024)
\end{botherref}
\endbibitem

%%% 108
\bibitem[\protect\citeauthoryear{Xu et~al.}{2024}]{xu2024dposurvey}
\begin{botherref}
\oauthor{\bsnm{Xu}, \binits{W.}},
\oauthor{\bsnm{Zhang}, \binits{Y.}},
\oauthor{\bsnm{Liu}, \binits{Y.}},
\oauthor{\bsnm{Hu}, \binits{Y.}},
\oauthor{\bsnm{Gao}, \binits{Y.}},
\oauthor{\bsnm{Luo}, \binits{X.}},
\oauthor{\bsnm{Li}, \binits{Z.}},
\oauthor{\bsnm{Zhang}, \binits{F.}},
\oauthor{\bsnm{Su}, \binits{H.}},
\oauthor{\bsnm{Zhu}, \binits{J.}}:
A comprehensive survey of direct preference optimization: Datasets, theories, variants, and applications.
arXiv preprint arXiv:2410.15595
(2024)
\end{botherref}
\endbibitem

%%% 109
\bibitem[\protect\citeauthoryear{Azar et~al.}{2024}]{azar2024general}
\begin{bchapter}
\bauthor{\bsnm{Azar}, \binits{M.G.}},
\bauthor{\bsnm{Guo}, \binits{Z.}},
\bauthor{\bsnm{Piot}, \binits{B.}},
\bauthor{\bsnm{Munos}, \binits{R.}},
\bauthor{\bsnm{Rowland}, \binits{M.}},
\bauthor{\bsnm{Valko}, \binits{M.}},
\bauthor{\bsnm{Calandriello}, \binits{D.}}:
\bctitle{A general theoretical paradigm to understand learning from human preferences}.
In: \bbtitle{Proceedings of the 27th International Conference on Artificial Intelligence and Statistics (AISTATS)}.
\bsertitle{Proceedings of Machine Learning Research},
vol. \bseriesno{238},
pp. \bfpage{4447}--\blpage{4455}
(\byear{2024})
\end{bchapter}
\endbibitem

%%% 110
\bibitem[\protect\citeauthoryear{Cao et~al.}{2024}]{cao2024beyond}
\begin{botherref}
\oauthor{\bsnm{Cao}, \binits{M.}},
\oauthor{\bsnm{Shu}, \binits{L.}},
\oauthor{\bsnm{Yu}, \binits{L.}},
\oauthor{\bsnm{Zhu}, \binits{Y.}},
\oauthor{\bsnm{Wichers}, \binits{N.}},
\oauthor{\bsnm{Liu}, \binits{Y.}},
\oauthor{\bsnm{Meng}, \binits{L.}}:
Beyond sparse rewards: Enhancing reinforcement learning with language model critique in text generation.
arXiv preprint arXiv:2401.07382
(2024)
\end{botherref}
\endbibitem

%%% 111
\bibitem[\protect\citeauthoryear{Chan et~al.}{2024}]{chan2024dense}
\begin{botherref}
\oauthor{\bsnm{Chan}, \binits{A.J.}},
\oauthor{\bsnm{Sun}, \binits{H.}},
\oauthor{\bsnm{Holt}, \binits{S.}},
\oauthor{\bsnm{Schaar}, \binits{M.}}:
Dense reward for free in reinforcement learning from human feedback.
arXiv preprint arXiv:2402.00782
(2024)
\end{botherref}
\endbibitem

%%% 112
\bibitem[\protect\citeauthoryear{Cheng et~al.}{2023}]{cheng2023adversarial}
\begin{botherref}
\oauthor{\bsnm{Cheng}, \binits{P.}},
\oauthor{\bsnm{Yang}, \binits{Y.}},
\oauthor{\bsnm{Li}, \binits{J.}},
\oauthor{\bsnm{Dai}, \binits{Y.}},
\oauthor{\bsnm{Hu}, \binits{T.}},
\oauthor{\bsnm{Cao}, \binits{P.}},
\oauthor{\bsnm{Du}, \binits{N.}},
\oauthor{\bsnm{Li}, \binits{X.}}:
Adversarial preference optimization: Enhancing your alignment via rm‑llm game.
arXiv preprint arXiv:2311.08045
(2023)
\end{botherref}
\endbibitem

%%% 113
\bibitem[\protect\citeauthoryear{Dai et~al.}{2023}]{dai2023safe}
\begin{botherref}
\oauthor{\bsnm{Dai}, \binits{J.}},
\oauthor{\bsnm{Pan}, \binits{X.}},
\oauthor{\bsnm{Sun}, \binits{R.}},
\oauthor{\bsnm{Ji}, \binits{J.}},
\oauthor{\bsnm{Xu}, \binits{X.}},
\oauthor{\bsnm{Liu}, \binits{M.}},
\oauthor{\bsnm{Wang}, \binits{Y.}},
\oauthor{\bsnm{Yang}, \binits{Y.}}:
Safe rlhf: Safe reinforcement learning from human feedback.
arXiv preprint arXiv:2310.12773
(2023)
\end{botherref}
\endbibitem

%%% 114
\bibitem[\protect\citeauthoryear{Dong et~al.}{2023}]{dong2023steerlm}
\begin{botherref}
\oauthor{\bsnm{Dong}, \binits{Y.}},
\oauthor{\bsnm{Wang}, \binits{Z.}},
\oauthor{\bsnm{Sreedhar}, \binits{M.N.}},
\oauthor{\bsnm{Wu}, \binits{X.}},
\oauthor{\bsnm{Kuchaiev}, \binits{O.}}:
Steerlm: Attribute conditioned sft as an (user-steerable) alternative to rlhf.
arXiv preprint arXiv:2310.05344
(2023)
\end{botherref}
\endbibitem

%%% 115
\bibitem[\protect\citeauthoryear{Ethayarajh et~al.}{2024}]{ethayarajh2024kto}
\begin{botherref}
\oauthor{\bsnm{Ethayarajh}, \binits{K.}},
\oauthor{\bsnm{Xu}, \binits{W.}},
\oauthor{\bsnm{Muennighoff}, \binits{N.}},
\oauthor{\bsnm{Jurafsky}, \binits{D.}},
\oauthor{\bsnm{Kiela}, \binits{D.}}:
Kto: Model alignment as prospect theoretic optimization.
arXiv preprint arXiv:2402.01306
(2024)
\end{botherref}
\endbibitem

%%% 116
\bibitem[\protect\citeauthoryear{Hong et~al.}{2024}]{hong2024orpo}
\begin{botherref}
\oauthor{\bsnm{Hong}, \binits{J.}},
\oauthor{\bsnm{Lee}, \binits{N.}},
\oauthor{\bsnm{Thorne}, \binits{J.}}:
{ORPO}: Monolithic preference optimization without reference model.
arXiv preprint arXiv:2403.07691
(2024)
\end{botherref}
\endbibitem

%%% 117
\bibitem[\protect\citeauthoryear{Kim et~al.}{2023}]{kim2023aligning}
\begin{botherref}
\oauthor{\bsnm{Kim}, \binits{S.}},
\oauthor{\bsnm{Bae}, \binits{S.}},
\oauthor{\bsnm{Shin}, \binits{J.}},
\oauthor{\bsnm{Kang}, \binits{S.}},
\oauthor{\bsnm{Kwak}, \binits{D.}},
\oauthor{\bsnm{Yoo}, \binits{K.M.}},
\oauthor{\bsnm{Seo}, \binits{M.}}:
Aligning large language models through synthetic feedback.
arXiv preprint arXiv:2305.13735
(2023)
\end{botherref}
\endbibitem

%%% 118
\bibitem[\protect\citeauthoryear{Lee et~al.}{2023}]{lee2023rlaif}
\begin{botherref}
\oauthor{\bsnm{Lee}, \binits{H.}},
\oauthor{\bsnm{Phatale}, \binits{S.}},
\oauthor{\bsnm{Mansoor}, \binits{H.}},
\oauthor{\bsnm{Lu}, \binits{K.R.}},
\oauthor{\bsnm{Mesnard}, \binits{T.}},
\oauthor{\bsnm{Ferret}, \binits{J.}},
\oauthor{\bsnm{Bishop}, \binits{C.}},
\oauthor{\bsnm{Hall}, \binits{E.}},
\oauthor{\bsnm{Carbune}, \binits{V.}},
\oauthor{\bsnm{Rastogi}, \binits{A.}}:
Rlaif: Scaling reinforcement learning from human feedback with ai feedback
(2023)
\end{botherref}
\endbibitem

%%% 119
\bibitem[\protect\citeauthoryear{Mahan et~al.}{2024}]{mahan2024generative}
\begin{botherref}
\oauthor{\bsnm{Mahan}, \binits{D.}},
\oauthor{\bsnm{Van~Phung}, \binits{D.}},
\oauthor{\bsnm{Rafailov}, \binits{R.}},
\oauthor{\bsnm{Blagden}, \binits{C.}},
\oauthor{\bsnm{Lile}, \binits{N.}},
\oauthor{\bsnm{Castricato}, \binits{L.}},
\oauthor{\bsnm{Franken}, \binits{J.-P.}},
\oauthor{\bsnm{Finn}, \binits{C.}},
\oauthor{\bsnm{Albalak}, \binits{A.}}:
Generative reward models.
arXiv preprint arXiv:2410.12832
(2024)
\end{botherref}
\endbibitem

%%% 120
\bibitem[\protect\citeauthoryear{Moskovitz et~al.}{2023}]{moskovitz2023confronting}
\begin{botherref}
\oauthor{\bsnm{Moskovitz}, \binits{T.}},
\oauthor{\bsnm{Singh}, \binits{A.K.}},
\oauthor{\bsnm{Strouse}, \binits{D.}},
\oauthor{\bsnm{Sandholm}, \binits{T.}},
\oauthor{\bsnm{Salakhutdinov}, \binits{R.}},
\oauthor{\bsnm{Dragan}, \binits{A.D.}},
\oauthor{\bsnm{McAleer}, \binits{S.}}:
Confronting reward model overoptimization with constrained rlhf.
arXiv preprint arXiv:2310.04373
(2023)
\end{botherref}
\endbibitem

%%% 121
\bibitem[\protect\citeauthoryear{Pang et~al.}{2024}]{pang2404iterative}
\begin{botherref}
\oauthor{\bsnm{Pang}, \binits{R.Y.}},
\oauthor{\bsnm{Yuan}, \binits{W.}},
\oauthor{\bsnm{Cho}, \binits{K.}},
\oauthor{\bsnm{He}, \binits{H.}},
\oauthor{\bsnm{Sukhbaatar}, \binits{S.}},
\oauthor{\bsnm{Weston}, \binits{J.}}:
Iterative reasoning preference optimization.
arXiv preprint arXiv:2404.19733
(2024)
\end{botherref}
\endbibitem

%%% 122
\bibitem[\protect\citeauthoryear{Park et~al.}{2024}]{park2024disentangling}
\begin{botherref}
\oauthor{\bsnm{Park}, \binits{R.}},
\oauthor{\bsnm{Rafailov}, \binits{R.}},
\oauthor{\bsnm{Ermon}, \binits{S.}},
\oauthor{\bsnm{Finn}, \binits{C.}}:
Disentangling length from quality in direct preference optimization.
arXiv preprint arXiv:2403.19159
(2024)
\end{botherref}
\endbibitem

%%% 123
\bibitem[\protect\citeauthoryear{Rafailov et~al.}{2024}]{rafailov2024direct}
\begin{botherref}
\oauthor{\bsnm{Rafailov}, \binits{R.}},
\oauthor{\bsnm{Sharma}, \binits{A.}},
\oauthor{\bsnm{Mitchell}, \binits{E.}},
\oauthor{\bsnm{Manning}, \binits{C.D.}},
\oauthor{\bsnm{Ermon}, \binits{S.}},
\oauthor{\bsnm{Finn}, \binits{C.}}:
Direct preference optimization: Your language model is secretly a reward model.
Advances in Neural Information Processing Systems
\textbf{36}
(2024)
\end{botherref}
\endbibitem

%%% 124
\bibitem[\protect\citeauthoryear{Rame et~al.}{2023}]{rame2023rewarded}
\begin{bchapter}
\bauthor{\bsnm{Rame}, \binits{A.}},
\bauthor{\bsnm{Couairon}, \binits{G.}},
\bauthor{\bsnm{Dancette}, \binits{C.}},
\bauthor{\bsnm{Gaya}, \binits{J.-B.}},
\bauthor{\bsnm{Shukor}, \binits{M.}},
\bauthor{\bsnm{Soulier}, \binits{L.}},
\bauthor{\bsnm{Cord}, \binits{M.}}:
\bctitle{Rewarded soups: Towards pareto-optimal alignment by interpolating weights fine-tuned on diverse rewards}.
In: \bbtitle{Advances in Neural Information Processing Systems (NeurIPS)},
vol. \bseriesno{36},
pp. \bfpage{71095}--\blpage{71134}
(\byear{2023})
\end{bchapter}
\endbibitem

%%% 125
\bibitem[\protect\citeauthoryear{Scheid et~al.}{2024}]{scheid2024optimal}
\begin{botherref}
\oauthor{\bsnm{Scheid}, \binits{A.}},
\oauthor{\bsnm{Boursier}, \binits{E.}},
\oauthor{\bsnm{Durmus}, \binits{A.}},
\oauthor{\bsnm{Jordan}, \binits{M.I.}},
\oauthor{\bsnm{Ménard}, \binits{P.}},
\oauthor{\bsnm{Moulines}, \binits{E.}},
\oauthor{\bsnm{Valko}, \binits{M.}}:
Optimal design for reward modeling in rlhf.
arXiv preprint arXiv:2410.17055
(2024)
\end{botherref}
\endbibitem

%%% 126
\bibitem[\protect\citeauthoryear{Wu et~al.}{2023}]{wu2023fine}
\begin{barticle}
\bauthor{\bsnm{Wu}, \binits{Z.}},
\bauthor{\bsnm{Hu}, \binits{Y.}},
\bauthor{\bsnm{Shi}, \binits{W.}},
\bauthor{\bsnm{Dziri}, \binits{N.}},
\bauthor{\bsnm{Suhr}, \binits{A.}},
\bauthor{\bsnm{Ammanabrolu}, \binits{P.}},
\bauthor{\bsnm{Smith}, \binits{N.A.}},
\bauthor{\bsnm{Ostendorf}, \binits{M.}},
\bauthor{\bsnm{Hajishirzi}, \binits{H.}}:
\batitle{Fine-grained human feedback gives better rewards for language model training}.
\bjtitle{Advances in Neural Information Processing Systems}
\bvolume{36},
\bfpage{59008}--\blpage{59033}
(\byear{2023})
\end{barticle}
\endbibitem

%%% 127
\bibitem[\protect\citeauthoryear{Xu et~al.}{2024}]{xu2024dpo}
\begin{botherref}
\oauthor{\bsnm{Xu}, \binits{S.}},
\oauthor{\bsnm{Fu}, \binits{W.}},
\oauthor{\bsnm{Gao}, \binits{J.}},
\oauthor{\bsnm{Ye}, \binits{W.}},
\oauthor{\bsnm{Liu}, \binits{W.}},
\oauthor{\bsnm{Mei}, \binits{Z.}},
\oauthor{\bsnm{Wang}, \binits{G.}},
\oauthor{\bsnm{Yu}, \binits{C.}},
\oauthor{\bsnm{Wu}, \binits{Y.}}:
Is dpo superior to ppo for llm alignment? a comprehensive study.
arXiv preprint arXiv:2404.10719
(2024)
\end{botherref}
\endbibitem

%%% 128
\bibitem[\protect\citeauthoryear{Yang et~al.}{2024}]{yang2024selective}
\begin{botherref}
\oauthor{\bsnm{Yang}, \binits{K.}},
\oauthor{\bsnm{Liu}, \binits{Z.}},
\oauthor{\bsnm{Xie}, \binits{Q.}},
\oauthor{\bsnm{Huang}, \binits{J.}},
\oauthor{\bsnm{Min}, \binits{E.}},
\oauthor{\bsnm{Ananiadou}, \binits{S.}}:
Selective preference optimization via token-level reward function estimation.
arXiv preprint arXiv:2408.13518
(2024)
\end{botherref}
\endbibitem

%%% 129
\bibitem[\protect\citeauthoryear{Yin et~al.}{2025}]{yin2025constrain}
\begin{bchapter}
\bauthor{\bsnm{Yin}, \binits{Q.}},
\bauthor{\bsnm{Leong}, \binits{C.T.}},
\bauthor{\bsnm{Zhang}, \binits{H.}},
\bauthor{\bsnm{Zhu}, \binits{M.}},
\bauthor{\bsnm{Yan}, \binits{H.}},
\bauthor{\bsnm{Zhang}, \binits{Q.}},
\bauthor{\bsnm{He}, \binits{Y.}},
\bauthor{\bsnm{Li}, \binits{W.}},
\bauthor{\bsnm{Wang}, \binits{J.}},
\bauthor{\bsnm{Zhang}, \binits{Y.}},
\bauthor{\bsnm{Yang}, \binits{L.}}:
\bctitle{Constrain alignment with sparse autoencoders}.
In: \bbtitle{Proceedings of the 2025 International Conference on Machine Learning (ICML)},
vol. \bseriesno{267}
(\byear{2025}).
\burl{https://arxiv.org/abs/2411.07618}
\end{bchapter}
\endbibitem

%%% 130
\bibitem[\protect\citeauthoryear{Zhang et~al.}{2024}]{zhang2024generative}
\begin{botherref}
\oauthor{\bsnm{Zhang}, \binits{L.}},
\oauthor{\bsnm{Hosseini}, \binits{A.}},
\oauthor{\bsnm{Bansal}, \binits{H.}},
\oauthor{\bsnm{Kazemi}, \binits{M.}},
\oauthor{\bsnm{Kumar}, \binits{A.}},
\oauthor{\bsnm{Agarwal}, \binits{R.}}:
Generative verifiers: Reward modeling as next-token prediction.
arXiv preprint arXiv:2408.15240
(2024)
\end{botherref}
\endbibitem

\end{thebibliography}

\end{document}